\documentclass{article} 
\usepackage{main,times}

\usepackage{amsmath,amsfonts,bm}

\def\eqref#1{equation~\ref{#1}}

\def\1{\bm{1}}

\DeclareMathAlphabet{\mathsfit}{\encodingdefault}{\sfdefault}{m}{sl}
\SetMathAlphabet{\mathsfit}{bold}{\encodingdefault}{\sfdefault}{bx}{n}

\usepackage{hyperref}
\usepackage{url}
\usepackage{algorithm}
\usepackage{algpseudocode}

\usepackage{colortbl}
\usepackage{array}
\definecolor{color3}{rgb}{0.95,0.95,0.95}
\definecolor{color4}{rgb}{0.90,0.9,0.9}
\usepackage{amsfonts}       
\usepackage{graphicx} 
\usepackage{caption} 
\usepackage{float}
\usepackage{xcolor}

\usepackage{booktabs}
\usepackage{multirow}
\usepackage{graphicx}
\usepackage[table]{xcolor} 
\usepackage{wrapfig}

\title{%
\vspace{-2em}
  \hspace{-1em}
  \raisebox{-2.5ex}{%
    \protect\includegraphics[height=5.0\fontcharht\font`\B]{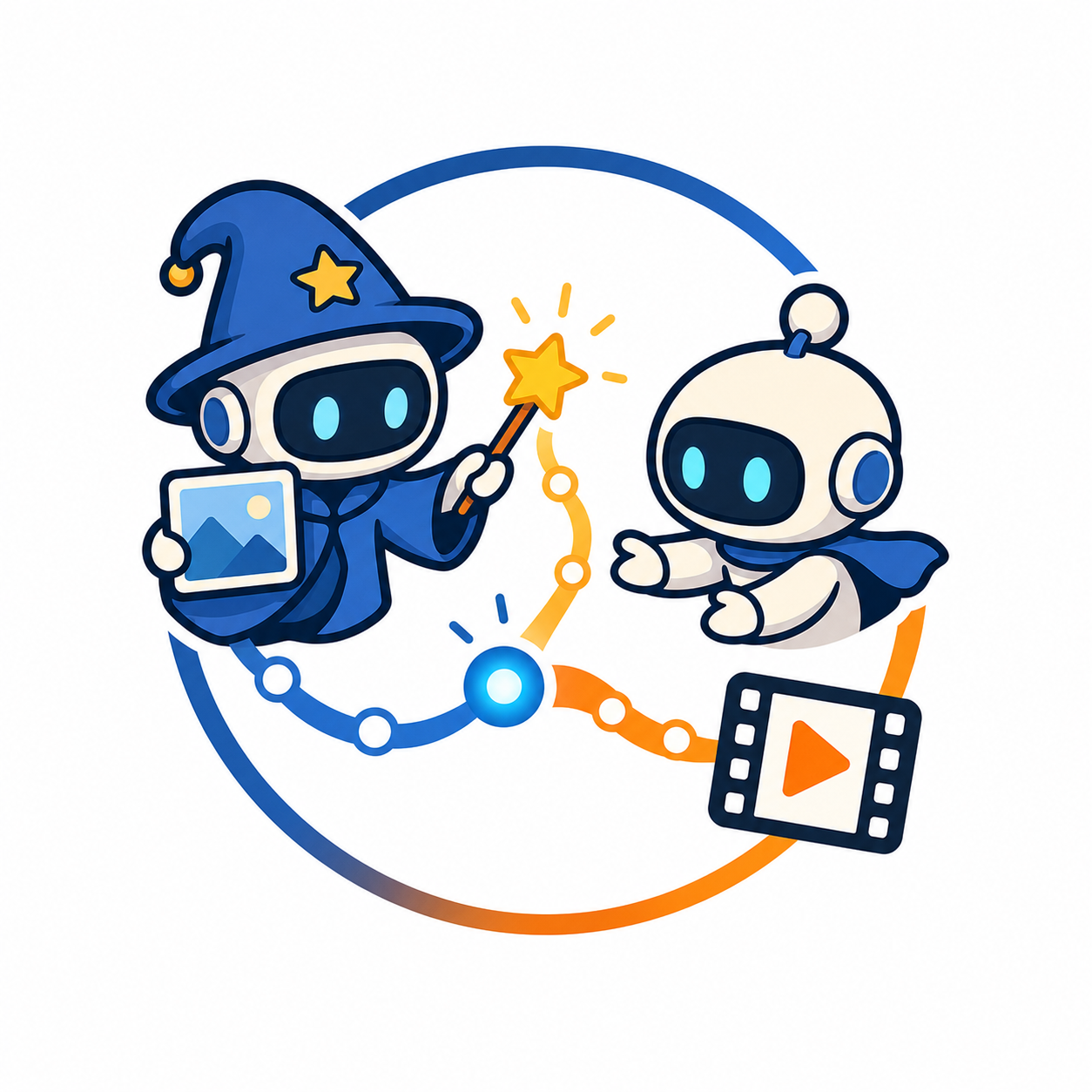}%
  }%
  HPSD: Hybrid-Policy Self-Distillation for Text-Image-to-Video Diffusion Models
}

\iclrfinalcopy

\author{Jiazi Bu$^{1,2,3*}$\quad
Pengyang Ling$^{4*\S}$\quad
Yujie Zhou$^{1,3*}$\quad
Yibin Wang$^{5,6}$\quad
Yuhang Zang$^{3}$ \\
\textbf{Xuanlang Dai}$^{5,3}$\quad
\textbf{Shengyuan Ding}$^{5,3}$ \quad
\textbf{Tianyi Wei$^{2}$}\quad
\textbf{Xiaohang Zhan$^{10}$}\quad
\textbf{Jiaqi Wang$^{6,9}$}\\
\textbf{Tong Wu$^{5}$}\quad
\textbf{Dahua Lin$^{3,7,8}$}\quad
\textbf{Xingang Pan$^{2\dag}$}\\
\small 
\textsuperscript{\rm 1}Shanghai Jiao Tong University \quad
\textsuperscript{\rm 2}S-Lab, Nanyang Technological University\quad
\textsuperscript{\rm 3}Shanghai AI Laboratory \\
\small
\textsuperscript{\rm 4}University of Science and Technology of China \quad
\textsuperscript{\rm 5}Fudan University \quad  
\textsuperscript{\rm 6}Shanghai Innovation Institute \\
\small
\textsuperscript{\rm 7}The Chinese University of Hong Kong \quad
\textsuperscript{\rm 8}CPII under InnoHK \quad 
\textsuperscript{\rm 9}JD.com \quad
\textsuperscript{\rm 10}Adobe Research \\
\url{https://bujiazi.github.io/hpsd.github.io/}
}

\begin{document}

{
  \renewcommand{\thefootnote}{\fnsymbol{footnote}}
  \footnotetext[1]{Equal contribution.  
  \textsuperscript{\S}Project leader.
  \textsuperscript{\dag}Corresponding author.}
}

\maketitle

\begin{figure}[h]
    \centering
    \vspace{-1em}
    \includegraphics[width=0.9\linewidth]{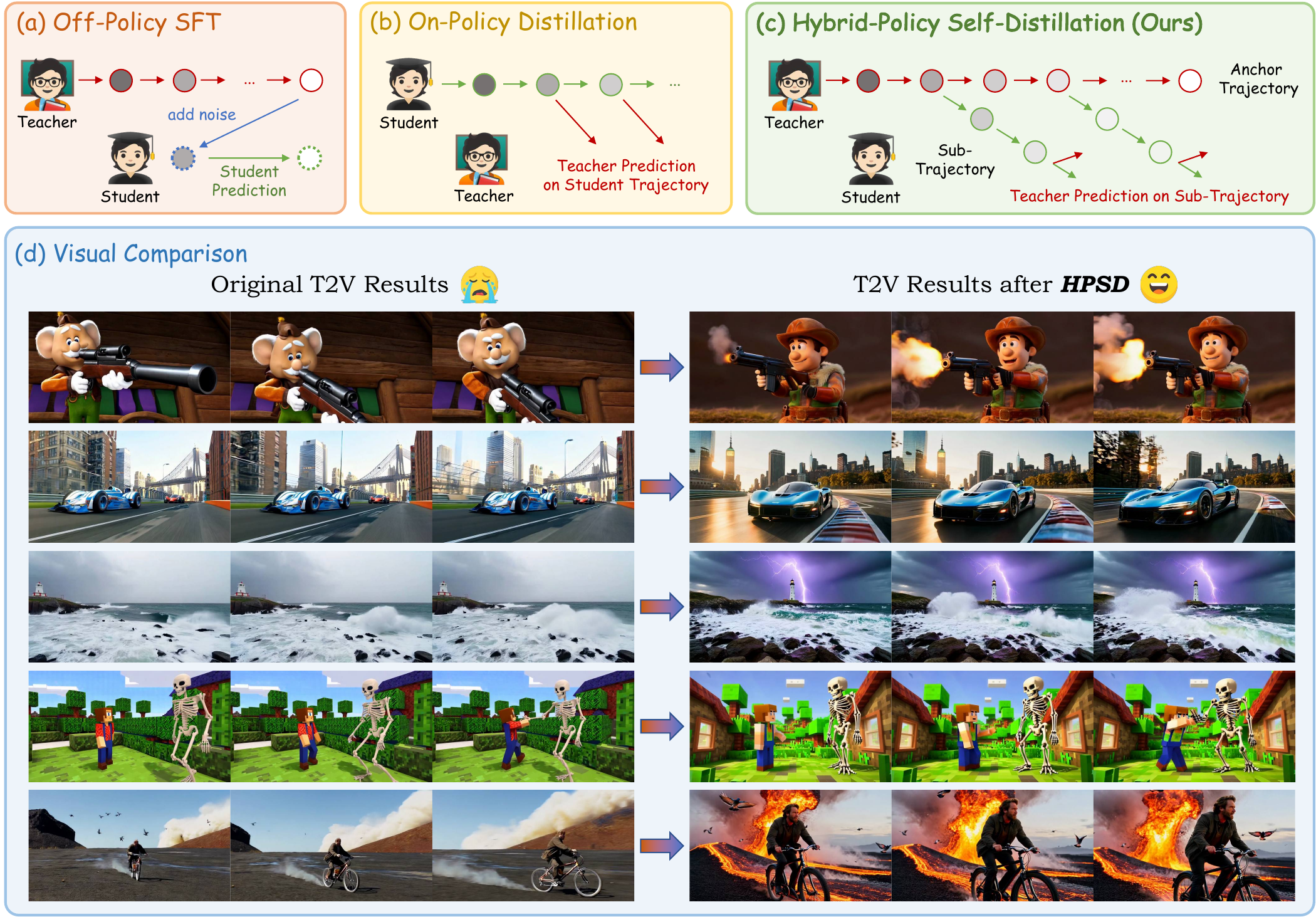}
    \caption{
        \textbf{Introduction to HPSD.} 
    (a) Off-policy SFT supervises on fixed teacher endpoints, which drift from the evolving student policy.
    (b) On-policy distillation queries the teacher at student-visited states, yet suffers from condition-state mismatch in TI2V models. 
    (c) HPSD anchors the student on the teacher trajectory and supervises roll-outs evolved under its own policy.
    (d) HPSD substantially boosts the base generation quality of TI2V models (WAN-2.2-TI2V-5B in this figure), especially in cinematic lighting, fine details, and balanced composition.
    \textbf{Prompts in the Appendix.}
    }
    \label{fig:teaser}
\end{figure}

\begin{abstract}
Text-Image-to-Video (TI2V) models are an emerging unified architecture, where a single model simultaneously supports text-to-video (T2V) and image-to-video (I2V) generation.
Given a high-quality first frame or a detailed textual prompt, TI2V models unlock substantially better visual quality than their T2V mode, raising a natural question: \textit{can the capability elicited by such privileged conditions be internalized into the model's own base generation ability?}
A common approach toward this goal is model self-distillation.
However, the most straightforward solution, supervised fine-tuning, follows an off-policy strategy: its supervision is confined to teacher-generated endpoints from a fixed offline distribution rather than student-visited states, 
lacking precise correction tailored to the evolving policy.
Recent on-policy distillation methods instead suffer from condition-state mismatch, where supervision is steered toward the given first frame instead of the student's actual content, 
misleading the denoising process.
To achieve self-distillation that absorbs the teacher's privileged prior while retaining precise policy correction, in this work, we propose \textbf{H}ybrid-\textbf{P}olicy \textbf{S}elf-\textbf{D}istillation (\textbf{HPSD}), a novel self-distillation framework where a single TI2V model acts as both teacher and student under different conditions: the teacher operates in TI2V mode with a high-quality first frame and an enhanced prompt, while the student runs in the base T2V mode with only the vanilla prompt.
Specifically, the student inherits off-policy teacher trajectory points as anchors, locally refines them toward its own policy, and finally receives velocity-level supervision on these self-generated roll-outs.
Extensive experiments demonstrate that HPSD significantly improves T2V performance while also delivering notable TI2V gains, effectively strengthening the model's base generation ability.
Our code will be released at \href{https://github.com/Bujiazi/HPSD}{HPSD Repo}.
\end{abstract}

\section{Introduction}

Recent years have witnessed remarkable progress in video generation, 
where diffusion/flow-based models~\citep{ho2020denoising,song2020denoising,song2020score,lipman2022flow,peebles2023scalable} 
have achieved a leap in high-fidelity video synthesis~\citep{wan2025wan, hacohen2024ltx, kong2024hunyuanvideo,bu2025bytheway,Zhou_2025_ICCV}. 
Building on these advances, the field has evolved into diverse generation paradigms,
among which text-to-video (T2V)~\citep{guo2023animatediff, kong2024hunyuanvideo, yang2025cogvideox}
and image-to-video (I2V)~\citep{blattmann2023stable, zhang2023i2vgen, xing2023dynamicrafter} are the most representative:
the former synthesizes videos from textual descriptions, while the latter animates a given reference image into a dynamic sequence. 
These two paradigms are increasingly unified within a single architecture, giving rise to Text-Image-to-Video (TI2V) models~\citep{lin2025stiv, wan2025wan, hacohen2026ltx,ma2026scaling}. 
Taking a textual prompt and an optional first-frame image as input, TI2V models produce faithful and coherent videos with a unified generative framework, 
allowing T2V and I2V to benefit from shared latent representations and motion priors while offering flexible conditioning for diverse user intents.

\begin{figure}[t]
    \centering
    \includegraphics[width=0.95\linewidth]{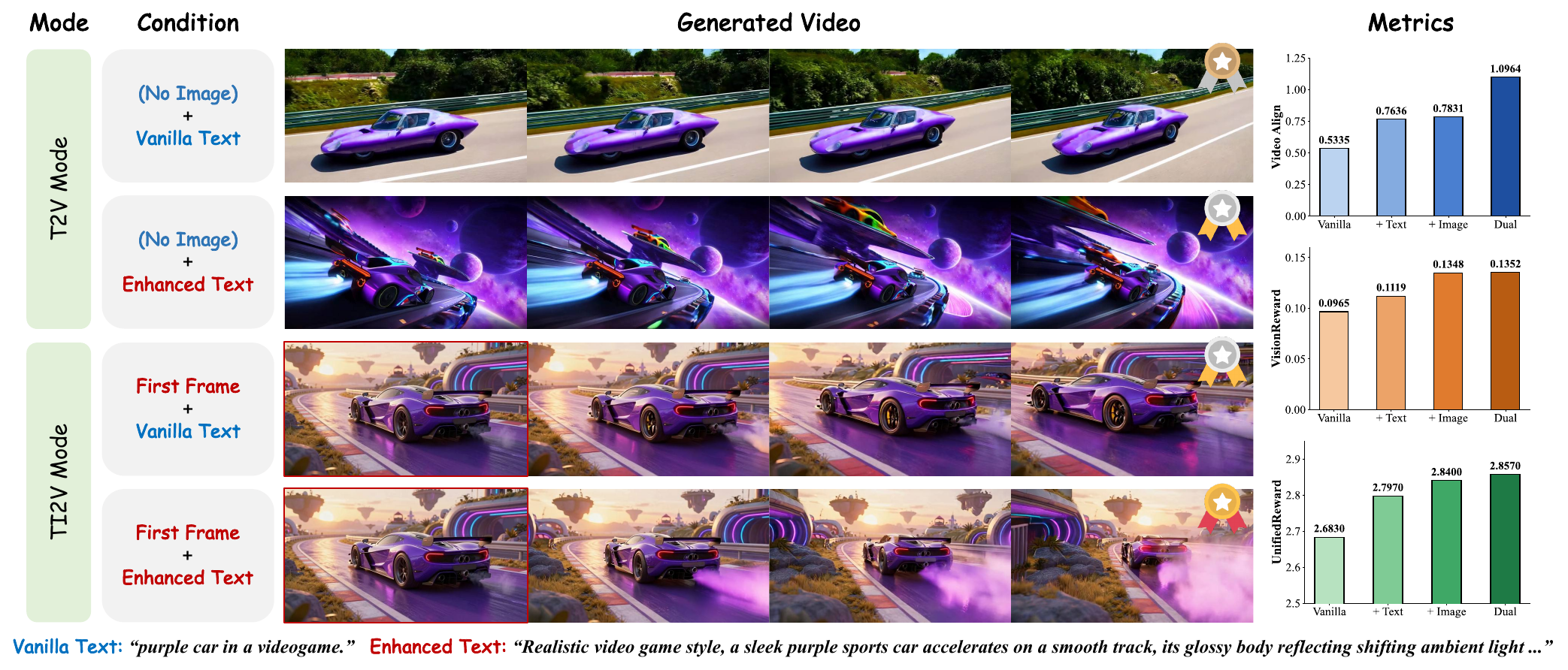}
    \vspace{-0.5em}
    \caption{
        \textbf{Condition-Elicited Capability.} 
        Conditioning a TI2V model on a high-quality first frame or an enhanced prompt clearly outperforms vanilla T2V. Their combination yields the best results across visuals and metrics, motivating HPSD's goal of internalizing this capability into the model.
        }
    \vspace{-1em}
    \label{fig:observation}
\end{figure}

In this work, we observe that within such a unified architecture, the TI2V mode conditioned on an additional high-quality first frame yields substantially better visual quality than the base T2V mode driven by a vanilla prompt, and that a detailed, well-crafted prompt alone brings considerable gains as well, as illustrated in Fig.~\ref{fig:observation}.
Moreover, these two conditions from different modalities are complementary: a carefully designed first frame combined with an enriched prompt further improves the generation quality beyond either alone.
We refer to such inputs as \textit{privileged conditions}, and to the underlying capability they awaken as the \textit{condition-elicited capability} of TI2V models.
Intuitively, privileged conditions inject additional content and motion priors into the denoising process, reducing generation uncertainty and improving video quality.
Notably, these quality differences arise from the same model operating under different external conditions, 
raising a natural question: \textit{can this condition-elicited capability be internalized into the model's own base generation ability?}

A common approach toward this goal is model self-distillation, 
which has long been studied in diffusion models~\citep{yin2024one,jiang2025no}. 
Nevertheless, the most straightforward solution, supervised fine-tuning (SFT) on teacher-generated videos, 
is inherently off-policy: the student is supervised only on teacher terminal outputs from a fixed offline distribution,
which drift away from the states it actually visits as training proceeds,  
precluding state-aware precise correction to the evolving policy, 
as shown in Fig.~\ref{fig:teaser} (a).
Recently, inspired by the success of on-policy distillation (OPD) in large language models (LLMs)~\citep{shenfeld2026self,yang2026self,zhao2026self,he2026self}, a growing line of research has brought this paradigm to diffusion and flow models~\citep{jiang2026dopsd,fang2026flow,li2026diffusionopd}, where the teacher is queried at student-visited states to provide dense, state-wise supervision along the student's own roll-outs, effectively mitigating the exposure bias of off-policy training, as depicted in Fig.~\ref{fig:teaser} (b).
In TI2V models, however, the privileged image condition is imposed as a fixed first frame that remains clean throughout denoising, whereas the student generates every frame 
without such conditioning.
The teacher is thus queried on mixed states where the prescribed first-frame content coexists with the student's own evolving content, and its supervision directs the denoising toward the former, conflicting with the latter---a failure we term \textit{condition-state mismatch} (Fig.~\ref{fig:mismatch}).
These limitations underscore the need for a new policy structure that absorbs the teacher's privileged prior while retaining precise policy correction.

To address these issues, we propose \textbf{H}ybrid-\textbf{P}olicy \textbf{S}elf-\textbf{D}istillation (\textbf{HPSD}), a novel self-distillation framework where a single TI2V model acts as both teacher and student under different conditions: the teacher operates in TI2V mode under the privileged image and prompt conditions, while the student runs in the base T2V mode with a vanilla text prompt.
The key idea is to let the student start from the teacher's trajectory and finish under its own policy, which proceeds in three steps:
(1) Before training, the privileged conditions are synthesized for each prompt with off-the-shelf generative models;
(2) During training, the teacher first rolls out its full denoising trajectory under privileged conditions, yielding an off-policy \textit{anchor trajectory} that carries the condition-elicited generation content;
(3) The student then continues denoising from intermediate states of the anchor trajectory with its own velocity field, and is supervised by the teacher on the resulting \textit{sub-trajectory}.
Since the supervised states are anchored by the teacher's policy yet evolved by the student's, we term this design a \textit{hybrid-policy} (Fig.~\ref{fig:teaser} (c)), which offers two core advantages: 
(i) the student inherits the teacher's states prescribed by the privileged conditions, so that the teacher's supervision no longer conflicts with the sample content; 
(ii) the supervised states arise from the student's own roll-out and thus remain aligned with its current policy.
By integrating off-policy anchoring with on-policy refinement, HPSD sidesteps both the exposure bias of off-policy training and the condition-state mismatch of on-policy distillation, offering a new principle for diffusion model distillation.
Extensive experiments on representative unified TI2V models (WAN-2.2~\citep{wan2025wan} and LTX-2.3~\citep{hacohen2026ltx}) 
demonstrate that HPSD surpasses both off-policy and on-policy baselines, 
substantially improving the T2V performance (Fig.~\ref{fig:teaser} (d)) while also delivering notable TI2V gains, 
indicating that the teacher and the student improve in tandem and that the model's base generation ability is consistently enhanced.

Our contributions are threefold:
(1) \textbf{Condition-Elicited Capability}: We identify that privileged conditions elicit stronger generation capability from unified TI2V models, and pose the new task of internalizing this capability into the model's base generation ability.
(2) \textbf{Hybrid-Policy Self-Distillation}: We reveal the failures of off-/on-policy distillation in this setting and propose HPSD, a hybrid-policy self-distillation framework that supervises the student on states anchored by the teacher's trajectory and evolved by its own policy.
(3) \textbf{Superior Performance}: Extensive experiments on two representative unified TI2V models validate that HPSD significantly strengthens the model's base generation ability beyond both off-/on-policy baselines.

\section{Related Work}

\textbf{Video Diffusion Models}. 
Video diffusion models have evolved from early U-Net-based architectures~\citep{guo2023animatediff, blattmann2023stable, zhang2023i2vgen} 
toward diffusion transformers (DiTs)~\citep{esser2024scaling}, 
giving rise to modern systems with remarkable visual fidelity and prompt adherence, such as OpenSora~\citep{zheng2024open}, 
CogVideoX~\citep{yang2025cogvideox}, LongCat-Video~\citep{longcatvideo}, HunyuanVideo series~\citep{kong2024hunyuanvideo,wu2025hunyuanvideo}, and WAN series~\citep{wan2025wan}.
These models generally follow two paradigms: text-to-video (T2V) models that synthesize videos from textual descriptions, and image-to-video (I2V) models that animate a given reference image. 
More recently, Text-Image-to-Video (TI2V) models~\citep{lin2025stiv,ma2026scaling} such as WAN-2.2~\citep{wan2025wan} and LTX-2~\citep{hacohen2026ltx} 
further unify the two paradigms within a single set of weights,
where an optional first-frame image is injected as a fixed clean condition into the denoising process.  
Beyond architecture design, a parallel line of research~\citep{wang2025promptenhancer,cheng2025vpo} improves generation quality by engineering richer input conditions at inference-time. 
In contrast, our HPSD paradigm internalizes the condition-elicited capabilities directly into the model's inherent base generation ability.

\textbf{Distillation for Diffusion Models}. 
Diffusion model distillation primarily focuses on step distillation~\citep{yin2024one, sauer2024adversarial, luo2023latent, yin2024improved}
and knowledge distillation~\citep{flux1-lite,yu2024representation}. 
To accelerate the sluggish sampling process, step distillation compresses a many-step teacher into a few-step student,
achieved either by imitating intermediate denoising transitions~\citep{luo2023latent,meng2023distillation,salimans2022progressive}
or aligning output distributions at specific timesteps~\citep{luo2025learning,yin2024one,yin2024improved}. 
Conversely, knowledge distillation transfers priors from a stronger teacher to improve the student's generation quality~\citep{fang2025tinyfusion,wu2026representation}. 
Unlike this line of study, we frame capability internalization as a self-distillation problem across different conditionings within the same model.

\textbf{On-Policy Distillation for Generative Models}. 
Initially grounded in large language models to mitigate exposure bias 
by matching distributions on student-generated sequences~\citep{agarwal2024policy,song2026survey,lu2025onpolicydistillation,jang2026stable}, 
on-policy distillation (OPD) has recently been adapted to diffusion 
and flow models to provide dense supervision along the student's own roll-outs. 
In this visual domain, existing methods primarily fall into two categories. 
The first is self-distillation, where a single model acts as both student and teacher; 
for instance, D-OPSD~\citep{jiang2026dopsd} and OPSD-V~\citep{liu2026opsd} 
guide the student using teachers conditioned on real images and temporal contexts, respectively. 
The second involves multi-teacher distillation, with works like DiffusionOPD~\citep{li2026diffusionopd}, 
Flow-OPD~\citep{fang2026flow}, and DanceOPD~\citep{zhou2026danceopd}, composing diverse capabilities
from multiple specialized teachers into a unified student via velocity matching or reward optimization. 
Distinct from these paradigms, our HPSD addresses the unique condition-state mismatch of OPD
in TI2V models through a novel hybrid-policy design.

\section{Method}

\subsection{Preliminaries}
\label{sec:prelim}

\textbf{Flow Matching}. Flow matching~\citep{liu2022flow, lipman2022flow} trains a time-dependent velocity field $v_\theta(x_t, t, c)$ that transports samples from a Gaussian prior to the data distribution. Given a data sample $x_0$ and noise $\epsilon \sim \mathcal{N}(0, I)$, the interpolated state and target velocity at time $t \in [0,1]$ are $x_t = (1-t)x_0 + t\epsilon$ and $v^* = \epsilon - x_0$, respectively. 
The model is optimized with $\mathcal{L}_{\text{FM}} = \mathbb{E}\|v_\theta(x_t, t, c) - v^*\|^2$. Sampling proceeds by integrating the learned velocity field from $t = 1$ to $0$:
\begin{equation}
\label{eq:ode}
\frac{\mathrm{d}x_t}{\mathrm{d}t} = v_\theta(x_t, t, c),
\end{equation}
where $c$ denotes input conditions such as textual prompts or reference images.

\textbf{TI2V Models}. A TI2V model~\citep{wan2025wan, hacohen2026ltx} supports both T2V and TI2V generation with a shared set of weights. In the T2V mode, the model is conditioned only on the text input, i.e., $v_\theta(x_t, t \mid c_\text{txt})$. In the TI2V mode, it is further conditioned on a given first-frame image $c_\text{img}$. Specifically, for TI2V generation, the first-frame condition is injected as a fixed clean latent and is excluded from the denoising process. For an $F$-frame video, the model is evaluated as:
\begin{equation}
\label{eq:ti2v}
v_\theta(\hat{x}_t, t \mid c_\text{txt}, c_\text{img}), \qquad \hat{x}_t = [\,c_\text{img},\, x_t^{(2:F)}\,],
\end{equation}
where the first-frame slot always contains the clean condition $c_\text{img}$, while the remaining frames are denoised at time $t$. 
\textit{Throughout, we use $x$ for T2V states whose frames share a single noise level, and $\hat{x}$ for TI2V states with a clean first-frame slot.}

\textbf{Distillation Paradigms}. Consider a teacher velocity field $\tilde{v}$ and a student $v_\phi$. Off-policy SFT fine-tunes the student on teacher-generated data. 
The teacher first completes a full denoising process to produce an endpoint $x_0^\text{Tea}$.
The student is supervised on its noised version $x_t^\text{Tea} = (1-t)x_0^\text{Tea} + t\epsilon$:
\begin{equation} 
\label{eq:off} 
\mathcal{L}_{\text{SFT}} = \mathbb{E}_{t}\,\big\|v_\phi(x_t^\text{Tea}, t) - v^*\big\|^2,
\end{equation} 
where $v^* = \epsilon - x_0^\text{Tea}$ is the analytic target velocity. 
For brevity, we suppress the conditional inputs here. 
On-policy distillation methods~\citep{jiang2026dopsd,zhou2026danceopd} instead supervise states produced by the current student. The student rolls out from its own policy, $x_t^\text{Stu} \sim \mathrm{Rollout}(v_\phi)$, and the teacher is evaluated at these student-visited states:
\begin{equation} 
\label{eq:on} 
\mathcal{L}_{\text{OPD}} = \mathbb{E}_{t}\,\big\|v_\phi(x_t^\text{Stu}, t) - \mathrm{sg}[\tilde{v}(x_t^\text{Stu}, t)]\big\|^2,
\end{equation}
where $\mathrm{sg}(\cdot)$ stands for the stop gradient operation.

\subsection{Observation and Analysis}
\label{sec:observation}
As observed in Fig.~\ref{fig:observation}, privileged conditions elicit a stronger generation capability from TI2V models: an enhanced prompt or a high-quality first frame alone already significantly improves over the base T2V mode, and their combination yields the best quality. To internalize this condition-elicited capability into the model's base generation ability, we formulate a self-distillation problem in which the teacher and the student are the \emph{same} TI2V model evaluated under different input conditions: the teacher runs under privileged conditions (an enhanced prompt $c_\text{txt}^{+}$ and a given first frame $c_\text{img}$), while the student operates in the base T2V mode with the original text prompt $c_\text{txt}$.

\begin{figure}[t]
    \centering
    \includegraphics[width=0.95\linewidth]{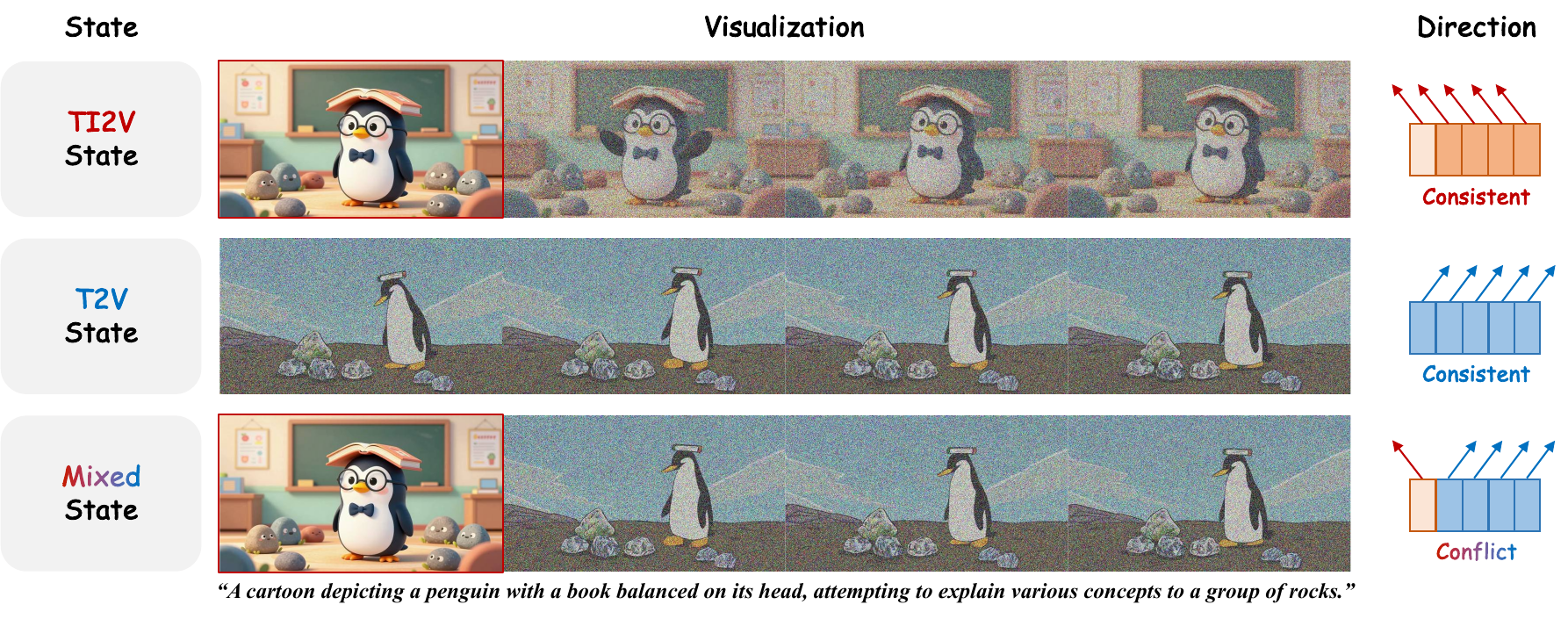}
    \vspace{-0.5em}
    \caption{
        \textbf{Condition-State Mismatch.} 
        While standard TI2V/T2V states maintain consistent content across frames, on-policy distillation creates an invalid mixed state by forcing a clean first frame alongside the student's unaligned T2V rollout, yielding conflicting, corrupted teacher supervision.
        }
    \vspace{-1em}
    \label{fig:mismatch}
\end{figure}

To be effective, this distillation must satisfy two competing requirements: providing state-aware precise correction to the evolving policy, while ensuring the teacher's supervision does not conflict with the student's actual sample content. 
While supervised fine-tuning (SFT) on teacher-generated videos successfully circumvents content conflicts, it remains fundamentally off-policy:
as the student policy evolves, the fixed offline supervision states drift away from the states the student actually visits. 
Conversely, recent on-policy distillation methods~\citep{jiang2026dopsd,zhou2026danceopd} evaluate at student-visited states to offer localized correction, but suffer from a severe \emph{condition-state mismatch} in the TI2V setting, as shown in Fig.~\ref{fig:mismatch}. 
Rolling out in the T2V mode, the student generates an internally coherent latent sequence $x_t^{\text{Stu},\,(1:F)}$.
However, since the TI2V teacher rigidly requires a clean first frame, querying it on a student's intermediate state creates a conflicted mixed input:
\begin{equation}
\label{eq:mismatch}
\hat{x}_t^{\text{Mismatch}} = [\,c_\text{img},\, x_t^{\text{Stu},\,(2:F)}\,].
\end{equation}
Here, the clean condition $c_\text{img}$ is forced to coexist with the student's partially denoised frames, which depict a completely different, free-running T2V content, causing a severe temporal discontinuity. 
Essentially, this unaligned mixture constitutes an invalid input state for the model. The teacher thus outputs a corrupted velocity field, generating contradictory signals that mislead the student's denoising process toward $c_\text{img}$, rather than refining the student's own content.
HPSD is structurally designed to resolve this dilemma and satisfy both requirements simultaneously.

\subsection{Hybrid-Policy Self-Distillation}
\label{sec:method}

\begin{figure}[t]
    \centering
    \includegraphics[width=0.9\linewidth]{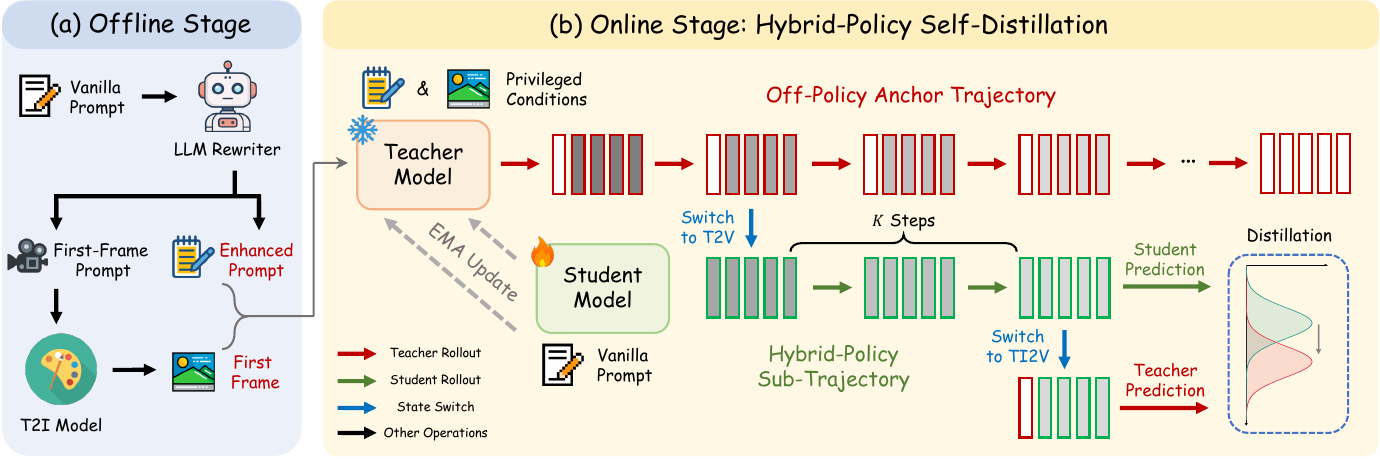}
    \vspace{-0.5em}
    \caption{
        \textbf{Overview of HPSD.} 
        (a) Offline Stage: Privileged conditions (enhanced prompt and first frame) are synthesized using auxiliary models.
        (b) Online Stage: The student evolves hybrid-policy sub-trajectories starting from the teacher's off-policy anchor states. Distillation on these states provides the student with precise policy correction anchored by the teacher's privileged content priors.
        \textit{Varying gray shading denotes the noise level over time, while solid white indicates clean frames.}
        }
    \vspace{-1em}
    \label{fig:pipeline}
\end{figure}

HPSD lets the student start from the teacher's trajectory and finish under its own policy, which proceeds in the following three steps. The overall framework of HPSD is illustrated in Fig.~\ref{fig:pipeline}.

\textbf{Privileged Condition Construction}.
Given a training prompt $c_\text{txt}$, we first synthesize the privileged conditions with off-the-shelf generative models: an external LLM rewrites $c_\text{txt}$ into an enhanced prompt $c_\text{txt}^{+}$ and designs a first-frame description $c_\text{ff}$, which an auxiliary text-to-image model then renders into the high-quality first frame $c_\text{img}$. All privileged conditions are precomputed offline and cached for training.
Formally, this process can be expressed as:
\begin{equation}
\label{eq:priv}
c_\text{txt}^{+},\, c_\text{ff} \sim \mathcal{M}_{\text{rw}}\big(\cdot \mid c_\text{txt}\big), \qquad c_\text{img} \sim \mathcal{G}_{\text{t2i}}\big(\cdot \mid c_\text{ff}\big),
\end{equation}
where $\mathcal{M}_{\text{rw}}$ denotes the LLM prompt rewriter, and $\mathcal{G}_{\text{t2i}}$ denotes the auxiliary text-to-image generator.

\textbf{Off-Policy Anchor Trajectory.}
The teacher first rolls out its full denoising trajectory under privileged conditions $(c_\text{txt}^{+}, c_\text{img})$, yielding an \emph{off-policy anchor trajectory} that carries the condition-elicited generation content:
\begin{equation}
\label{eq:anchor_traj}
\hat{x}_{t_{j+1}}^{\text{Tea},\,(2:F)} = \hat{x}_{t_j}^{\text{Tea},\,(2:F)} + \big(t_{j+1} - t_j\big)\, \tilde{v}\big(\hat{x}_{t_j}^\text{Tea}, t_j \,\big|\, c_\text{txt}^{+}, c_\text{img}\big), \qquad \hat{x}_{1}^{\text{Tea},\,(2:F)} = \epsilon, \;\; \epsilon \sim \mathcal{N}(0, I).
\end{equation}
 From it we collect intermediate states $\{\hat{x}_{t_i}^\text{Tea}\}_{i=1}^N$ at anchor steps $\mathcal{A} = \{t_i\}_{i=1}^N$ spread along the trajectory. Each anchor state $\hat{x}_{t_i}^\text{Tea}$ is then converted into a \emph{student-compatible} state $x_{t_i}^\text{Tea}$: since the teacher's first-frame slot always holds the clean condition (Eq.~\ref{eq:ti2v}) rather than a denoising state at time $t_i$, we re-noise it to the current noise level,
\begin{equation}
\label{eq:anchor}
x_{t_i}^{\text{Tea}} = \big[\,(1-{t_i})\,c_\text{img} + {t_i}\,\epsilon,\; \hat{x}_{t_i}^{\text{Tea},\,(2:F)}\,\big], \qquad \epsilon \sim \mathcal{N}(0, I),
\end{equation}
so that all frames share the noise level ${t_i}$, matching what the student sees during T2V inference. 
This reconciles the teacher's privileged content with the student's input format, turning each anchor state into a valid starting point of the student's own roll-out.

\textbf{Hybrid-Policy Sub-Trajectory.}
From each start point $x^{\text{Tea}}_{t_i}$, the student denoises $K$ steps forward with its own velocity field,
\begin{equation}
\label{eq:sub_traj}
x_{t_{j+1}}^\text{Hyb} = x_{t_j}^\text{Hyb} + \big(t_{j+1} - t_j\big)\, v_\phi\big(x_{t_j}^\text{Hyb}, t_j \,\big|\, c_\text{txt}\big), \qquad j = i, \dots, i{+}K{-}1, \;\; x_{t_i}^\text{Hyb} := x_{t_i}^\text{Tea},
\end{equation}
yielding a short \emph{hybrid-policy sub-trajectory} whose terminal state $x_{t_{i+K}}^\text{Hyb}$ inherits the teacher's content yet is evolved by the student's current policy. To supervise at this state, the teacher must be queried in its own input format: following Eq.~\ref{eq:ti2v}, we re-impose the clean first frame onto $x_{t_{i+K}}^\text{Hyb}$,
\begin{equation}
\label{eq:hybrid_query}
\hat{x}_{t_{i+K}}^\text{Hyb} = \big[\,c_\text{img},\, x_{t_{i+K}}^{\text{Hyb},\,(2:F)}\,\big],
\end{equation}
and evaluate the teacher at $\hat{x}_{t_{i+K}}^\text{Hyb}$. The student is then supervised on the frames being denoised:
\begin{equation}
\label{eq:hpsd}
\mathcal{L}_{\text{HPSD}} = \mathbb{E}_{t_i \sim \mathcal{A}}\;\Big\|\, v_\phi\big(x_{t_{i+K}}^\text{Hyb}, t_{i+K} \,\big|\, c_\text{txt}\big) - \mathrm{sg}\Big[\tilde{v}\big(\hat{x}_{t_{i+K}}^\text{Hyb}, t_{i+K} \,\big|\, c_\text{txt}^{+}, c_\text{img}\big)\Big] \Big\|^{2}_{(2:F)}.
\end{equation}
The sub-trajectory length $K$ interpolates between the two paradigms: $K = 0$ reduces to off-policy distillation on the teacher's trajectory, while a larger $K$ approaches on-policy supervision. 
Anchored by the teacher's trajectory and evolved by the student's policy, these intermediate states bridge both paradigms, which is why we denote this design as \emph{hybrid-policy}.
Note that the loss is applied from the second frame onward, as the teacher's first-frame velocity carries no denoising meaning under the clean-condition input format.

\textbf{HPSD Training.}
Alg.~\ref{alg:hpsd} outlines the complete HPSD pipeline.
In the offline stage, we synthesize and cache the privileged conditions $(c_\text{txt}^{+}, c_\text{img})$ for every training prompt.
During online training, the process iterates over prompt batches. 
First, the teacher rolls out an anchor trajectory to provide student-compatible anchor states. 
Next, the student denoises for $K$ steps from these anchors to form hybrid-policy sub-trajectories, receiving teacher supervision at the terminal states. 
The teacher is updated via an exponential moving average (EMA) of the student~\citep{jiang2026dopsd}, which stabilizes the target distribution and ensures that both roles improve in tandem.

\begin{figure}[t] 
\centering
\resizebox{0.8\linewidth}{!}{ 
\begin{minipage}{\linewidth} 
\begin{algorithm}[H] 
\caption{Hybrid-Policy Self-Distillation (HPSD)}
\label{alg:hpsd}
\begin{algorithmic}[1]
\Require TI2V model with student $v_\phi$ and EMA teacher $\tilde{v}$; prompt set $\mathcal{D}$; sub-trajectory length $K$; anchor steps $\mathcal{A}=\{t_i\}_{i=1}^N$
\Statex \textcolor{gray}{\//\// \textit{Offline stage: privileged-condition construction}}
\ForAll{$c_\text{txt} \in \mathcal{D}$}
\State Enhance the prompt $c_\text{txt} \to c_\text{txt}^{+}$ and synthesize the first frame $c_\text{img}$
\Comment{Eq.~\ref{eq:priv}}
\EndFor
\Statex \textcolor{gray}{\//\// \textit{Online stage: hybrid-policy self-distillation}}
\For{each training step}
\State Sample a prompt batch $\{c_\text{txt}\} \subset \mathcal{D}$; load cached $(c_\text{txt}^{+}, c_\text{img})$
\Statex \textcolor{gray}{\quad\//\// \textit{Off-policy anchor trajectory}}
\State Roll out the teacher $\tilde{v}(\cdot \mid c_\text{txt}^{+}, c_\text{img})$; collect anchor states $\{\hat{x}_{t_i}^\text{Tea}\}$
\Comment{Eq.~\ref{eq:anchor_traj}}
\State Convert each anchor into a student-compatible state $x_{t_i}^\text{Tea}$ \Comment{Eq.~\ref{eq:anchor}}
\Statex \textcolor{gray}{\quad\//\// \textit{Hybrid-policy sub-trajectory}}
\ForAll{anchors $x_{t_i}^\text{Tea}$}
\State Denoise $K$ steps with the student $v_\phi(\cdot \mid c_\text{txt})$ $\to x_{t_{i+K}}^\text{Hyb}$
\Comment{Eq.~\ref{eq:sub_traj}}
\State Re-impose the clean first frame to build the teacher query $\hat{x}_{t_{i+K}}^\text{Hyb}$ \Comment{Eq.~\ref{eq:hybrid_query}}
\EndFor
\Statex \textcolor{gray}{\quad\//\// \textit{Velocity-level matching}}
\State Compute $\mathcal{L}_{\text{HPSD}}$ and update the student $v_\phi$
\Comment{Eq.~\ref{eq:hpsd}}
\State Update the teacher $\tilde{v}$ with the EMA of the student
\EndFor
\end{algorithmic}
\end{algorithm}
\end{minipage}
}
\vspace{-1em}
\end{figure}

\section{Experiments} 

\subsection{Implementation Details}

\textbf{Training Datasets}. 
We adopt the video training dataset used in Pref-GRPO~\citep{Pref-GRPO&UniGenBench}
as the prompt set for HPSD and all baselines,
which consists of $\sim$50K prompts providing broad coverage across diverse themes and subject categories.
For privileged condition construction, we employ Qwen3.6-27B~\citep{qwen36_27b} as the LLM rewriter 
and Z-Image-Turbo~\citep{team2025zimage} as the high-quality first-frame generator.
Further details and additional experiments with an alternative first-frame generator are deferred to Section~\ref{sec:details} and Section~\ref{sec:add-quantitative} in the Appendix, respectively.

\textbf{Base Models}. 
Experiments are conducted on two leading TI2V models: WAN-2.2-TI2V-5B~\citep{wan2025wan} and LTX-2.3~\citep{hacohen2026ltx}. 
Both models are trained at their native resolutions and inference steps: 1280$\times$704 with 50 steps for WAN-2.2, and 768$\times$512 with 30 steps for LTX-2.3. 
For LTX-2.3, we only consider its first-stage generation and do not adopt the second-stage super-resolution and refinement.
For efficiency, both models use 33 frames during training.

\textbf{Baselines for Comparison}.
The compared methods encompass: 
(i) the base models in their T2V mode~\citep{wan2025wan,hacohen2026ltx};
(ii) Off-policy Supervised Fine-Tuning with flow-matching loss~\citep{lipman2022flow};
(iii) On-Policy Distillation;
(iv) D-OPSD~\citep{jiang2026dopsd}, a recent self-distillation method for diffusion models.
Note that D-OPSD requires a multimodal encoder to inject image information into the denoising process, 
which is typically unavailable in TI2V models; we therefore only apply D-OPSD to distill the textual privileged condition.

\textbf{Training Details}. 
Unless specified otherwise, all experiments share the following setup. 
We train for 500 steps on 8 H200 GPUs with a batch size of 1 per GPU, using the AdamW optimizer with a learning rate of $1\times10^{-4}$ and \texttt{bfloat16} (bf16) mixed-precision training for efficiency. 
The anchor steps $\mathcal{A}$ are randomly sampled from the anchor trajectory, while the sub-trajectory length $K$ is set to 3 and the EMA decay rate is set to 0.999.
All methods use LoRA with $r=32, \alpha=64$.

\textbf{Evaluation Details}. 
The evaluation set consists of 500 diverse video prompts sampled from the VideoDPO~\citep{liu2024videodpo} and VideoFeedback~\citep{he2024videoscore} datasets.
For a thorough evaluation, a diverse set of metrics is employed: 
(i) Video Reward Models, including VideoAlign~\citep{liu2025improving}, 
VisionReward~\citep{xu2026visionreward}, and UnifiedReward-v1/v2 (UR-v1/v2, unified image-video reward models)~\citep{unifiedreward}; 
(ii) Image Reward Models, where we adopt frame-wise HPS Score~\citep{wu2023human} 
and frame-wise CLIP Score~\citep{radford2021learning} to assess per-frame quality; 
and (iii) VBench~\citep{huang2023vbench} for comprehensive video generation evaluation.

\begin{table}[t]
\centering
\caption{
Quantitative comparison of different methods on various TI2V backbones. 
The best results are in \textbf{bold}, while the second-best result is \underline{underlined}.
UR-v2-A, UR-v2-P and UR-v2-S represent the Alignment, Physics and Style dimensions of UnifiedReward-v2, respectively.
}
\vspace{-1em}
\label{tab:reward_results}
\resizebox{0.9\linewidth}{!}{
\begin{tabular}{lccccccccc}
\toprule
\textbf{Backbone} & \textbf{Method} & \textbf{VideoAlign} & \textbf{VisionReward} & \textbf{UR-v1} & \textbf{UR-v2-A} & \textbf{UR-v2-P} & \textbf{UR-v2-S} & \textbf{HPS} & \textbf{CLIP} \\ \midrule
\multirow{5}{*}{\textbf{WAN-2.2}}
& Vanilla T2V & 0.5335 & 0.0965 & 2.683 & 2.802 & 3.167 & 3.100 & 0.2472 & 0.3684 \\
& On-Policy Distillation & 0.2613 & 0.0482 & 2.463 & 2.724 & 3.171 & 3.204 & 0.2561 & 0.2847 \\
& D-OPSD (\textit{Text}) & 0.6379 & \underline{0.1076} & 2.739 & 2.825 & \underline{3.189} & 3.187 & 0.2493 & \underline{0.3712} \\
& Supervised Fine-Tuning & \underline{1.2046} & 0.1043 & \underline{2.763} & \underline{2.854} & 3.181 & \underline{3.207} & \underline{0.2670} & 0.3710 \\
& \cellcolor{color3}{\textbf{HPSD (Ours)}} & \cellcolor{color3}{\textbf{1.8753}} & \cellcolor{color3}{\textbf{0.1191}} & \cellcolor{color3}{\textbf{2.812}} & \cellcolor{color3}{\textbf{2.890}} & \cellcolor{color3}{\textbf{3.203}} & \cellcolor{color3}{\textbf{3.275}} & \cellcolor{color3}{\textbf{0.2815}} & \cellcolor{color3}{\textbf{0.3765}} \\
\midrule
\multirow{5}{*}{\textbf{LTX-2.3}}
& Vanilla T2V & 0.2307 & 0.0566 & 2.648 & 2.877 & 3.100 & 2.956 & 0.2068 & 0.3320 \\
& On-Policy Distillation & 0.0242 & 0.0405 & 2.594 & 2.825 & 3.089 & 3.093 & 0.2278 & 0.3048 \\
& D-OPSD (\textit{Text}) & 0.4380 & \underline{0.1008} & \underline{2.824} & \textbf{2.903} & \underline{3.173} & 3.114 & \underline{0.2451} & \underline{0.3808} \\
& Supervised Fine-Tuning & \underline{0.9584} & 0.0826 & 2.733 & 2.873 & 3.068 & \underline{3.131} & 0.2349 & 0.3626 \\
& \cellcolor{color3}{\textbf{HPSD (Ours)}} & \cellcolor{color3}{\textbf{1.5244}} & \cellcolor{color3}{\textbf{0.1123}} & \cellcolor{color3}{\textbf{2.827}} & \cellcolor{color3}{\underline{2.887}} & \cellcolor{color3}{\textbf{3.198}} & \cellcolor{color3}{\textbf{3.182}} & \cellcolor{color3}{\textbf{0.2731}} & \cellcolor{color3}{\textbf{0.3853}} \\
\bottomrule
\end{tabular}
}
\end{table}
\begin{figure}[t]
    \centering
    \includegraphics[width=0.9\linewidth]{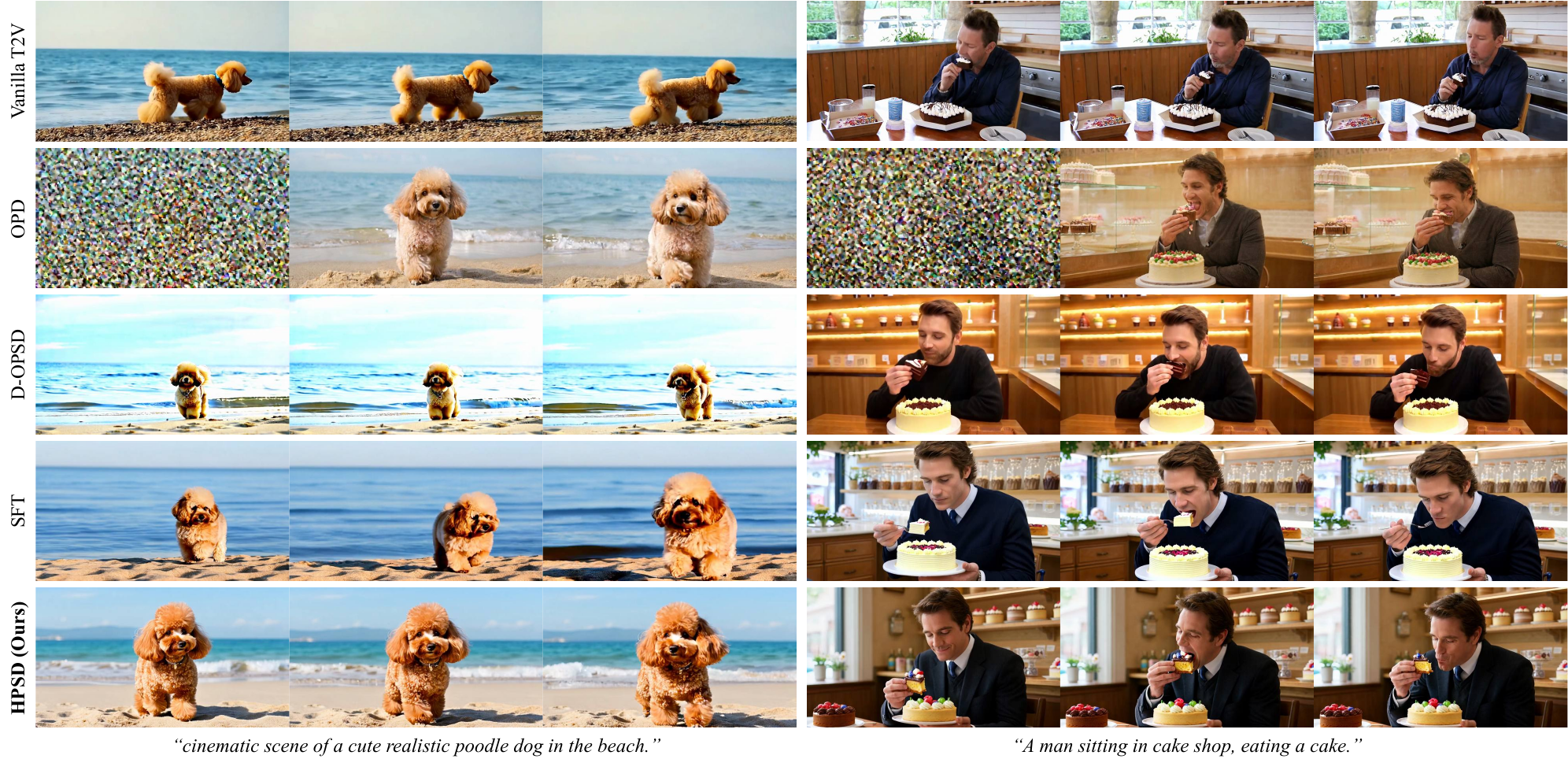}
    \vspace{-1em}
    \caption{
        \textbf{Qualitative Comparisons with Baselines on WAN-2.2.} 
        Best viewed zoomed in.
        }
    \label{fig:wan22}
    \vspace{-1em}
\end{figure}

\subsection{Main Results}

\begin{figure}[!t]
    \centering
    \includegraphics[width=0.9\linewidth]{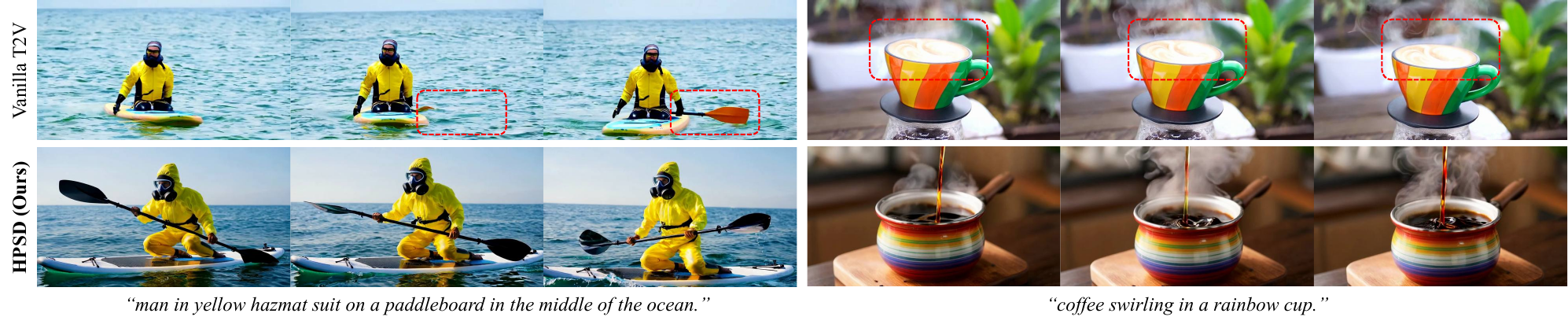}
    \vspace{-1em}
    \caption{
        \textbf{Improved Consistency \& Physical Plausibility.} 
        (Left) Vanilla T2V exhibits limited temporal consistency (paddle appears abruptly), while HPSD ensures coherent subject retention. (Right) Vanilla T2V lacks physical realism (static liquid), whereas HPSD renders natural fluid dynamics.
        }
    \vspace{-1em}
    \label{fig:consistency_physics}
\end{figure}

\textbf{Quantitative Evaluation}.
The quantitative results are presented in Tab.~\ref{tab:reward_results} and Tab.~\ref{tab:vbench_results}.
For reward metrics (Tab.~\ref{tab:reward_results}), 
HPSD consistently outperforms baselines, topping all eight reward metrics on WAN-2.2 and seven on LTX-2.3, notably boosting VideoAlign by $\sim$50\% over the second-best SFT.
Conversely, standard OPD severely degrades T2V performance due to its condition-state mismatch in training, leading to early-frame blurriness and content incoherence.
While D-OPSD offers some gains, its encoder's modality constraints preclude distilling image conditions, capping its upper bound.
On VBench (Tab.~\ref{tab:vbench_results}), HPSD leads on most dimensions for both models.
Although the Dynamic Degree is slightly lower, we argue that it does not imply degraded video quality; rather, we observe significantly enhanced temporal consistency, fewer abrupt changes, and better physical plausibility (also evidenced by the improved UR-v2-Physics score in Tab.~\ref{tab:reward_results}), as shown in Fig.~\ref{fig:consistency_physics}.

\begin{figure}[t]
    \centering
    \includegraphics[width=0.9\linewidth]{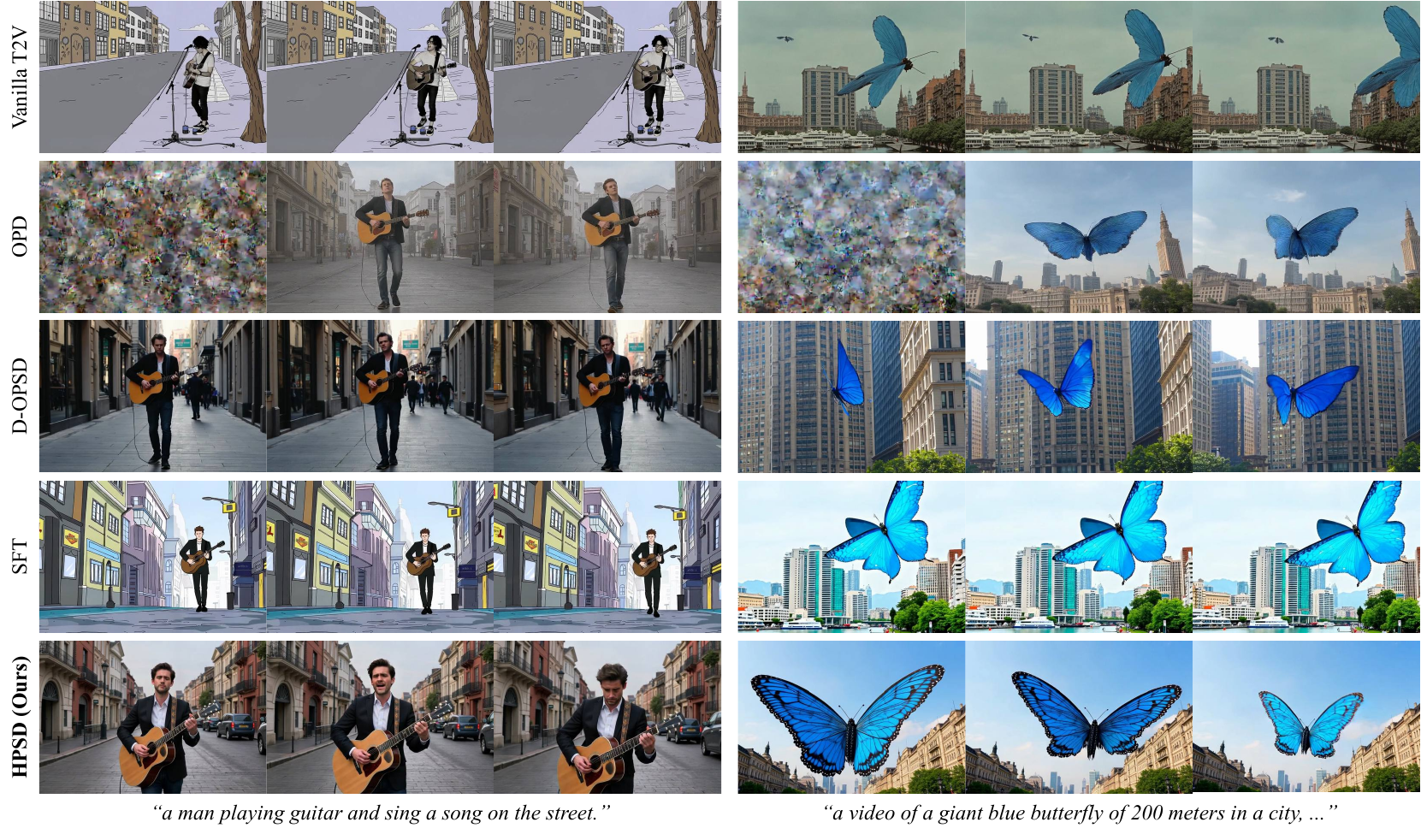}
    \vspace{-1em}
    \caption{
        \textbf{Qualitative Comparisons with Baselines on LTX-2.3.} 
        Best viewed zoomed in.
        }
    \vspace{-1em}
    \label{fig:ltx23}
\end{figure}

\textbf{Qualitative Comparison}. 
Visual comparisons in Fig.~\ref{fig:wan22} and Fig.~\ref{fig:ltx23} corroborate the metrics.
Vanilla T2V outputs are often desaturated with coarse textures, whereas On-Policy Distillation collapses entirely (e.g., noisy early frames and ghosting artifacts), stemming from the aforementioned condition-state mismatch.
While SFT and D-OPSD produce plausible content, they still lack fine details, consistent subjects, and cinematic lighting.
In contrast, HPSD delivers significantly sharper textures, richer colors, and superior prompt adherence (e.g., the cinematic poodle and vividly textured butterfly), visually confirming the successful internalization of condition-elicited capabilities.
Additional visual comparison results are presented in Section~\ref{sec:add-qualitative} in the Appendix.

\begin{table}[t]
\centering
\caption{
Quantitative results on VBench metrics. 
The best results are in \textbf{bold}, while the second-best result is \underline{underlined}.
SC: Subject Consistency; BC: Background Consistency; MS: Motion Smoothness; DD: Dynamic Degree; AQ: Aesthetic Quality; IQ: Imaging Quality.
}
\vspace{-1em}
\label{tab:vbench_results}
\resizebox{0.9\linewidth}{!}{
\begin{tabular}{lccccccc}
\toprule
\textbf{Backbone} & \textbf{Method} & \textbf{VBench-SC} & \textbf{VBench-BC} & \textbf{VBench-MS} & \textbf{VBench-DD} & \textbf{VBench-AQ} & \textbf{VBench-IQ} \\ \midrule
\multirow{5}{*}{\textbf{WAN-2.2}}
& Vanilla T2V & 0.9654 & 0.9667 & 0.9878 & \textbf{0.560} & 0.5773 & 0.6981 \\
& On-Policy Distillation & 0.5241 & 0.7284 & 0.9799 & 0.516 & 0.6066 & 0.6683 \\
& D-OPSD (\textit{Text}) & \underline{0.9691} & \underline{0.9692} & \underline{0.9874} & 0.404 & 0.5889 & 0.6300 \\
& Supervised Fine-Tuning & 0.9679 & 0.9675 & 0.9837 & \underline{0.526} & \underline{0.6095} & \textbf{0.7049} \\
& \cellcolor{color3}{\textbf{HPSD (Ours)}} & \cellcolor{color3}{\textbf{0.9722}} & \cellcolor{color3}{\textbf{0.9705}} & \cellcolor{color3}{\textbf{0.9880}} & \cellcolor{color3}{0.422} & \cellcolor{color3}{\textbf{0.6343}} & \cellcolor{color3}{\underline{0.6996}} \\
\midrule
\multirow{5}{*}{\textbf{LTX-2.3}}
& Vanilla T2V & 0.9388 & \underline{0.9467} & 0.9901 & \underline{0.330} & 0.5791 & 0.6517 \\
& On-Policy Distillation & 0.5337 & 0.7715 & 0.9899 & \textbf{0.366} & 0.5800 & 0.6345 \\
& D-OPSD (\textit{Text}) & 0.9419 & 0.9454 & 0.9897 & 0.236 & \underline{0.5994} & 0.6383 \\
& Supervised Fine-Tuning & \underline{0.9466} & 0.9465 & \underline{0.9903} & 0.278 & \underline{0.5994} & \underline{0.6723} \\
& \cellcolor{color3}{\textbf{HPSD (Ours)}} & \cellcolor{color3}{\textbf{0.9515}} & \cellcolor{color3}{\textbf{0.9554}} & \cellcolor{color3}{\textbf{0.9905}} & \cellcolor{color3}{0.302} & \cellcolor{color3}{\textbf{0.6375}} & \cellcolor{color3}{\textbf{0.6886}} \\
\bottomrule
\end{tabular}
}
\end{table}

\subsection{Ablation Study}

The ablation studies are conducted on the WAN-2.2 model, with the results presented in Tab.~\ref{tab:ablation}.

\textbf{Effects of Sub-Trajectory Length $K$}. 
The sub-trajectory length $K$ acts as a tunable knob interpolating between off-policy and on-policy paradigms.
As shown in Tab.~\ref{tab:ablation} (a), HPSD's performance initially improves but later deteriorates as $K$ increases, 
suggesting that an excessively short sub-trajectory remains overly off-policy, failing to reach the student's actual distribution, whereas a prolonged $K$ pushes the framework toward pure on-policy distillation, drifting too far from the teacher's prior and exacerbating the condition-state mismatch (Fig.~\ref{fig:mismatch}).
Therefore, $K = 3$ is chosen to balance the teacher's condition-elicited anchoring with the student's policy-aligned correction.

\textbf{Effects of Privileged Conditions}. 
Tab.~\ref{tab:ablation} (b) reveals a clear step-wise improvement when scaling the teacher's privileged conditions.
Compared to Vanilla T2V, distilling only the privileged image ($+c_\text{img}$) brings substantial gains, while combining both privileges ($+c_\text{txt}^{+}+c_\text{img}$) achieves the peak performance. 
This validates that richer conditions unlock stronger teacher capabilities, yielding higher-quality guidance for internalization. 
Notably, the text-only ablation is omitted, as text conditions do not trigger the condition-state mismatch and are readily handled by standard OPD.

\begin{figure}[!t]
    \centering
    \includegraphics[width=0.9\linewidth]{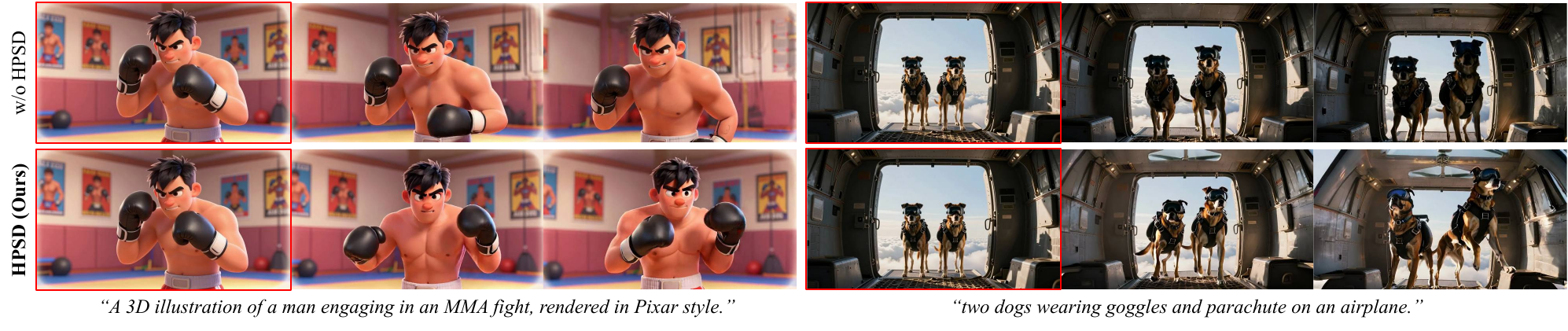}
    \vspace{-1em}
    \caption{
        \textbf{Effects of HPSD on TI2V mode.} 
        The \textcolor{red}{red} boxes indicate the shared first frames.
        }
    \label{fig:ti2v}
\end{figure}

\textbf{Effects of HPSD on TI2V Mode}.
Although optimized for T2V generation, HPSD simultaneously enhances the model's TI2V performance (Tab.~\ref{tab:ablation} (c)), improving VideoAlign by $\sim$55\%.
These gains stem from our asymmetric prompt distillation: 
forcing the student using a vanilla prompt to align with the teacher using an enhanced prompt intrinsically trains the model to better interpret and execute plain textual inputs, regardless of first frame conditioning. 
Consequently, HPSD yields richer physical interactions and superior prompt adherence (as illustrated in Fig.~\ref{fig:ti2v}), confirming 
an intrinsic enhancement of the model's shared generation priors.

\begin{table}[t]
\centering
\caption{Ablation experiments on HPSD components and hyperparameters. 
}
\vspace{-1em}
\label{tab:ablation}
\resizebox{0.9\linewidth}{!}{
\begin{tabular}{lccccccc}
\toprule
\textbf{Components} & \textbf{Values \& Choices} & \textbf{VideoAlign} & \textbf{VisionReward} & \textbf{UR-v1} & \textbf{UR-v2-A} & \textbf{UR-v2-P} & \textbf{UR-v2-S} \\ \midrule
\multirow{5}{*}{{(a) Sub-Trajectory Length $K$}}
& $K=0$ & 1.5587 & 0.1131 & 2.750 & 2.853 & 3.163 & 3.223 \\
& $K=1$ & 1.6653 & 0.1144 & 2.761 & 2.846 & 3.180 & 3.232 \\
& \cellcolor{color3}{\textbf{$\mathbf{K=3}$}} & \cellcolor{color3}{\textbf{1.8753}} & \cellcolor{color3}{\underline{0.1191}} & \cellcolor{color3}{\textbf{2.812}} & \cellcolor{color3}{\textbf{2.890}} & \cellcolor{color3}{\textbf{3.203}} & \cellcolor{color3}{\textbf{3.275}} \\
& $K=5$ & \underline{1.7794} & \textbf{0.1194} & \underline{2.790} & 2.865 & \underline{3.190} & \underline{3.260} \\
& $K=7$ & 1.7130 & 0.1168 & 2.777 & \underline{2.869} & 3.189 & 3.251 \\
\midrule
\multirow{3}{*}{{(b) Privileged Conditions $c_\text{txt}^{+}$ \& $c_\text{img}$}}
& Vanilla T2V & 0.5335 & 0.0965 & 2.683 & 2.802 & 3.167 & 3.100 \\
& $+c_\text{img}$ & \underline{1.5006} & \underline{0.1157} & \underline{2.787} & \underline{2.855} & \underline{3.194} & \underline{3.247} \\
& \cellcolor{color3}{\textbf{$\mathbf{+c_\text{txt}^{+}+c_\text{img}}$ (HPSD)}} & \cellcolor{color3}{\textbf{1.8753}} & \cellcolor{color3}{\textbf{0.1191}} & \cellcolor{color3}{\textbf{2.812}} & \cellcolor{color3}{\textbf{2.890}} & \cellcolor{color3}{\textbf{3.203}} & \cellcolor{color3}{\textbf{3.275}} \\
\midrule
\multirow{2}{*}{{(c) Effect on TI2V Mode}}
& Vanilla TI2V & 0.7831 & \textbf{0.1348} & 2.840 & 2.926 & 3.211 & \textbf{3.273} \\
& \cellcolor{color3}{\textbf{TI2V after HPSD}} & \cellcolor{color3}{\textbf{1.2139}} & \cellcolor{color3}{0.1344} & \cellcolor{color3}{\textbf{2.859}} & \cellcolor{color3}{\textbf{2.932}} & \cellcolor{color3}{\textbf{3.215}} & \cellcolor{color3}{3.271} \\
\bottomrule
\end{tabular}
}
\end{table}

\section{Limitation and Discussion}
While HPSD significantly enhances the base T2V generation capability, it introduces certain computational overheads. 
First, constructing privileged conditions requires querying auxiliary LLM and T2I models, which incurs extra data synthesis costs. 
However, this overhead is a one-time offline process and can be effectively mitigated by deploying distilled or quantized generative models (e.g., Z-Image-Turbo used in our experiments). 
Moreover, the performance ceiling of HPSD can naturally scale with the continuous advancements of these auxiliary models.
Second, during training, rolling out the student to form the hybrid-policy sub-trajectory requires additional forward passes compared to SFT and OPD. 
Nevertheless, these intermediate steps operate strictly in inference mode and do not require storing gradients, thereby avoiding substantial memory overhead.

\section{Conclusion}
We propose Hybrid-Policy Self-Distillation (HPSD) to internalize the superior, condition-elicited capabilities of TI2V models into their base generation ability. To resolve the lack of state-aware precise correction in off-policy methods and the condition-state mismatch in on-policy approaches, HPSD supervises the student on sub-trajectories anchored by the teacher's privileged states but evolved under its own policy. Experiments on WAN-2.2 and LTX-2.3 demonstrate that HPSD significantly boosts base T2V performance and concurrently enhances the TI2V mode, effectively strengthening the model's inherent generation ability and offering a novel distillation paradigm.

\newpage

\bibliographystyle{main}
\bibliography{main}

@article{wu2023human,
  title={Human preference score v2: A solid benchmark for evaluating human preferences of text-to-image synthesis},
  author={Wu, Xiaoshi and Hao, Yiming and Sun, Keqiang and Chen, Yixiong and Zhu, Feng and Zhao, Rui and Li, Hongsheng},
  journal={arXiv preprint arXiv:2306.09341},
  year={2023}
}

@inproceedings{radford2021learning,
  title={Learning transferable visual models from natural language supervision},
  author={Radford, Alec and Kim, Jong Wook and Hallacy, Chris and Ramesh, Aditya and Goh, Gabriel and Agarwal, Sandhini and Sastry, Girish and Askell, Amanda and Mishkin, Pamela and Clark, Jack and others},
  booktitle={International conference on machine learning},
  pages={8748--8763},
  year={2021},
  organization={PmLR}
}

@article{ho2020denoising,
  title={Denoising diffusion probabilistic models},
  author={Ho, Jonathan and Jain, Ajay and Abbeel, Pieter},
  journal={Advances in neural information processing systems},
  volume={33},
  pages={6840--6851},
  year={2020}
}

@article{song2020denoising,
  title={Denoising diffusion implicit models},
  author={Song, Jiaming and Meng, Chenlin and Ermon, Stefano},
  journal={arXiv preprint arXiv:2010.02502},
  year={2020}
}

@article{song2020score,
  title={Score-based generative modeling through stochastic differential equations},
  author={Song, Yang and Sohl-Dickstein, Jascha and Kingma, Diederik P and Kumar, Abhishek and Ermon, Stefano and Poole, Ben},
  journal={arXiv preprint arXiv:2011.13456},
  year={2020}
}

@inproceedings{peebles2023scalable,
  title={Scalable diffusion models with transformers},
  author={Peebles, William and Xie, Saining},
  booktitle={Proceedings of the IEEE/CVF international conference on computer vision},
  pages={4195--4205},
  year={2023}
}

@inproceedings{esser2024scaling,
  title={Scaling rectified flow transformers for high-resolution image synthesis},
  author={Esser, Patrick and Kulal, Sumith and Blattmann, Andreas and Entezari, Rahim and M{\"u}ller, Jonas and Saini, Harry and Levi, Yam and Lorenz, Dominik and Sauer, Axel and Boesel, Frederic and others},
  booktitle={Forty-first international conference on machine learning},
  year={2024}
}

@article{kong2024hunyuanvideo,
  title={Hunyuanvideo: A systematic framework for large video generative models},
  author={Kong, Weijie and Tian, Qi and Zhang, Zijian and Min, Rox and Dai, Zuozhuo and Zhou, Jin and Xiong, Jiangfeng and Li, Xin and Wu, Bo and Zhang, Jianwei and others},
  journal={arXiv preprint arXiv:2412.03603},
  year={2024}
}

@article{wan2025wan,
  title={Wan: Open and advanced large-scale video generative models},
  author={Wan, Team and Wang, Ang and Ai, Baole and Wen, Bin and Mao, Chaojie and Xie, Chen-Wei and Chen, Di and Yu, Feiwu and Zhao, Haiming and Yang, Jianxiao and others},
  journal={arXiv preprint arXiv:2503.20314},
  year={2025}
}

@article{guo2023animatediff,
  title={Animatediff: Animate your personalized text-to-image diffusion models without specific tuning},
  author={Guo, Yuwei and Yang, Ceyuan and Rao, Anyi and Liang, Zhengyang and Wang, Yaohui and Qiao, Yu and Agrawala, Maneesh and Lin, Dahua and Dai, Bo},
  journal={arXiv preprint arXiv:2307.04725},
  year={2023}
}

@article{lipman2022flow,
  title={Flow matching for generative modeling},
  author={Lipman, Yaron and Chen, Ricky TQ and Ben-Hamu, Heli and Nickel, Maximilian and Le, Matt},
  journal={arXiv preprint arXiv:2210.02747},
  year={2022}
}

@article{liu2022flow,
  title={Flow straight and fast: Learning to generate and transfer data with rectified flow},
  author={Liu, Xingchao and Gong, Chengyue and Liu, Qiang},
  journal={arXiv preprint arXiv:2209.03003},
  year={2022}
}

@article{blattmann2023stable,
  title={Stable video diffusion: Scaling latent video diffusion models to large datasets},
  author={Blattmann, Andreas and Dockhorn, Tim and Kulal, Sumith and Mendelevitch, Daniel and Kilian, Maciej and Lorenz, Dominik and Levi, Yam and English, Zion and Voleti, Vikram and Letts, Adam and others},
  journal={arXiv preprint arXiv:2311.15127},
  year={2023}
}

@article{Pref-GRPO&UniGenBench,
  title={Pref-GRPO: Pairwise Preference Reward-based GRPO for Stable Text-to-Image Reinforcement Learning},
  author={Wang, Yibin and Li, Zhimin and Zang, Yuhang and Zhou, Yujie and Bu, Jiazi and Wang, Chunyu and Lu, Qinglin and Jin, Cheng and Wang, Jiaqi},
  journal={arXiv preprint arXiv:2508.20751},
  year={2025}
}

@InProceedings{Zhou_2025_ICCV,
    author    = {Zhou, Yujie and Bu, Jiazi and Ling, Pengyang and Zhang, Pan and Wu, Tong and Huang, Qidong and Li, Jinsong and Dong, Xiaoyi and Zang, Yuhang and Cao, Yuhang and Rao, Anyi and Wang, Jiaqi and Niu, Li},
    title     = {Light-A-Video: Training-free Video Relighting via Progressive Light Fusion},
    booktitle = {Proceedings of the IEEE/CVF International Conference on Computer Vision (ICCV)},
    month     = {October},
    year      = {2025},
    pages     = {13315-13325}
}

@inproceedings{bu2025bytheway,
  title={ByTheWay: Boost Your Text-to-Video Generation Model to Higher Quality in a Training-free Way},
  author={Bu, Jiazi and Ling, Pengyang and Zhang, Pan and Wu, Tong and Dong, Xiaoyi and Zang, Yuhang and Cao, Yuhang and Lin, Dahua and Wang, Jiaqi},
  booktitle={Proceedings of the Computer Vision and Pattern Recognition Conference},
  pages={12999--13008},
  year={2025}
}

@article{hacohen2026ltx,
  title={LTX-2: Efficient Joint Audio-Visual Foundation Model},
  author={HaCohen, Yoav and Brazowski, Benny and Chiprut, Nisan and Bitterman, Yaki and Kvochko, Andrew and Berkowitz, Avishai and Shalem, Daniel and Lifschitz, Daphna and Moshe, Dudu and Porat, Eitan and others},
  journal={arXiv preprint arXiv:2601.03233},
  year={2026}
}

@article{hacohen2024ltx,
  title={Ltx-video: Realtime video latent diffusion},
  author={HaCohen, Yoav and Chiprut, Nisan and Brazowski, Benny and Shalem, Daniel and Moshe, Dudu and Richardson, Eitan and Levin, Eran and Shiran, Guy and Zabari, Nir and Gordon, Ori and others},
  journal={arXiv preprint arXiv:2501.00103},
  year={2024}
}

@misc{longcatvideo,
      title={LongCat-Video Technical Report}, 
      author={Meituan LongCat Team and Xunliang Cai and Qilong Huang and Zhuoliang Kang and Hongyu Li and Shijun Liang and Liya Ma and Siyu Ren and Xiaoming Wei and Rixu Xie and Tong Zhang},
      year={2025},
      eprint={2510.22200},
      archivePrefix={arXiv},
      primaryClass={cs.CV},
      url={https://arxiv.org/abs/2510.22200}, 
}

@inproceedings{yang2025cogvideox,
  title={Cogvideox: Text-to-video diffusion models with an expert transformer},
  author={Yang, Zhuoyi and Teng, Jiayan and Zheng, Wendi and Ding, Ming and Huang, Shiyu and Xu, Jiazheng and Yang, Yuanming and Hong, Wenyi and Zhang, Xiaohan and Feng, Guanyu and others},
  booktitle={International Conference on Learning Representations},
  volume={2025},
  pages={83048--83077},
  year={2025}
}

@article{zhang2023i2vgen,
  title={I2vgen-xl: High-quality image-to-video synthesis via cascaded diffusion models},
  author={Zhang, Shiwei and Wang, Jiayu and Zhang, Yingya and Zhao, Kang and Yuan, Hangjie and Qin, Zhiwu and Wang, Xiang and Zhao, Deli and Zhou, Jingren},
  journal={arXiv preprint arXiv:2311.04145},
  year={2023}
}

@article{xing2023dynamicrafter,
  title={DynamiCrafter: Animating Open-domain Images with Video Diffusion Priors},
  author={Xing, Jinbo and Xia, Menghan and Zhang, Yong and Chen, Haoxin and Yu, Wangbo and Liu, Hanyuan and Wang, Xintao and Wong, Tien-Tsin and Shan, Ying},
  journal={arXiv preprint arXiv:2310.12190},
  year={2023}
}

@article{zheng2024open,
  title={Open-sora: Democratizing efficient video production for all},
  author={Zheng, Zangwei and Peng, Xiangyu and Yang, Tianji and Shen, Chenhui and Li, Shenggui and Liu, Hongxin and Zhou, Yukun and Li, Tianyi and You, Yang},
  journal={arXiv preprint arXiv:2412.20404},
  year={2024}
}

@article{team2025zimage,
  title={Z-Image: An Efficient Image Generation Foundation Model with Single-Stream Diffusion Transformer},
  author={Z-Image Team},
  journal={arXiv preprint arXiv:2511.22699},
  year={2025}
}

@article{jiang2026dopsd,
      title={D-OPSD: On-Policy Self-Distillation for Continuously Tuning Step-Distilled Diffusion Models},
      author={Jiang, Dengyang and Jin, Xin and Liu, Dongyang and Wang, Zanyi and Zheng, Mingzhe and Du, Ruoyi and Yang, Xiangpeng and Wu, Qilong and Li, Zhen and Gao, Peng and Yang, Harry and Hoi, Steven},
      journal={arXiv preprint arXiv:2605.05204},
      year={2026}
}

@article{liu2025improving,
  title={Improving video generation with human feedback},
  author={Liu, Jie and Liu, Gongye and Liang, Jiajun and Yuan, Ziyang and Liu, Xiaokun and Zheng, Mingwu and Wu, Xiele and Wang, Qiulin and Qin, Wenyu and Xia, Menghan and others},
  journal={arXiv preprint arXiv:2501.13918},
  year={2025}
}

@article{liu2024videodpo,
  title={Videodpo: Omni-preference alignment for video diffusion generation},
  author={Liu, Runtao and Wu, Haoyu and Ziqiang, Zheng and Wei, Chen and He, Yingqing and Pi, Renjie and Chen, Qifeng},
  journal={arXiv preprint arXiv:2412.14167},
  year={2024}
}

@article{he2024videoscore,
  title = {VideoScore: Building Automatic Metrics to Simulate Fine-grained Human Feedback for Video Generation},
  author = {He, Xuan and Jiang, Dongfu and Zhang, Ge and Ku, Max and Soni, Achint and Siu, Sherman and Chen, Haonan and Chandra, Abhranil and Jiang, Ziyan and Arulraj, Aaran and Wang, Kai and Do, Quy Duc and Ni, Yuansheng and Lyu, Bohan and Narsupalli, Yaswanth and Fan, Rongqi and Lyu, Zhiheng and Lin, Yuchen and Chen, Wenhu},
  journal = {ArXiv},
  year = {2024},
  volume={abs/2406.15252},
  url = {https://arxiv.org/abs/2406.15252},
}

@inproceedings{xu2026visionreward,
  title={Visionreward: Fine-grained multi-dimensional human preference learning for image and video generation},
  author={Xu, Jiazheng and Huang, Yu and Cheng, Jiale and Yang, Yuanming and Xu, Jiajun and Wang, Yuan and Duan, Wenbo and Yang, Shen and Jin, Qunlin and Li, Shurun and others},
  booktitle={Proceedings of the AAAI Conference on Artificial Intelligence},
  volume={40},
  number={13},
  pages={11269--11277},
  year={2026}
}

@article{unifiedreward,
  title={Unified reward model for multimodal understanding and generation},
  author={Wang, Yibin and Zang, Yuhang and Li, Hao and Jin, Cheng and Wang, Jiaqi},
  journal={arXiv preprint arXiv:2503.05236},
  year={2025}
}

@InProceedings{huang2023vbench,
 title={{VBench}: Comprehensive Benchmark Suite for Video Generative Models},
 author={Huang, Ziqi and He, Yinan and Yu, Jiashuo and Zhang, Fan and Si, Chenyang and Jiang, Yuming and Zhang, Yuanhan and Wu, Tianxing and Jin, Qingyang and Chanpaisit, Nattapol and Wang, Yaohui and Chen, Xinyuan and Wang, Limin and Lin, Dahua and Qiao, Yu and Liu, Ziwei},
 booktitle={Proceedings of the IEEE/CVF Conference on Computer Vision and Pattern Recognition},
 year={2024}
}

@inproceedings{yin2024one,
  title={One-step diffusion with distribution matching distillation},
  author={Yin, Tianwei and Gharbi, Micha{\"e}l and Zhang, Richard and Shechtman, Eli and Durand, Fredo and Freeman, William T and Park, Taesung},
  booktitle={Proceedings of the IEEE/CVF conference on computer vision and pattern recognition},
  pages={6613--6623},
  year={2024}
}

@inproceedings{sauer2024adversarial,
  title={Adversarial diffusion distillation},
  author={Sauer, Axel and Lorenz, Dominik and Blattmann, Andreas and Rombach, Robin},
  booktitle={European Conference on Computer Vision},
  pages={87--103},
  year={2024},
  organization={Springer}
}

@inproceedings{luo2025learning,
  title={Learning few-step diffusion models by trajectory distribution matching},
  author={Luo, Yihong and Hu, Tianyang and Sun, Jiacheng and Cai, Yujun and Tang, Jing},
  booktitle={Proceedings of the IEEE/CVF International Conference on Computer Vision},
  pages={17719--17728},
  year={2025}
}

@inproceedings{yin2024improved,
  title={Improved distribution matching distillation for fast image synthesis},
  author={Yin, Tianwei and Gharbi, Micha{\"e}l and Park, Taesung and Zhang, Richard and Shechtman, Eli and Durand, Fredo and Freeman, William T},
  booktitle={The Thirty-eighth Annual Conference on Neural Information Processing Systems},
  year={2024}
}

@article{luo2023latent,
  title={Latent consistency models: Synthesizing high-resolution images with few-step inference},
  author={Luo, Simian and Tan, Yiqin and Huang, Longbo and Li, Jian and Zhao, Hang},
  journal={arXiv preprint arXiv:2310.04378},
  year={2023}
}

@inproceedings{agarwal2024policy,
  title={On-policy distillation of language models: Learning from self-generated mistakes},
  author={Agarwal, Rishabh and Vieillard, Nino and Zhou, Yongchao and Stanczyk, Piotr and Ramos Garea, Sabela and Geist, Matthieu and Bachem, Olivier},
  booktitle={International Conference on Learning Representations},
  volume={2024},
  pages={21246--21263},
  year={2024}
}

@article{song2026survey,
  title={A survey of on-policy distillation for large language models},
  author={Song, Mingyang and Zheng, Mao},
  journal={arXiv preprint arXiv:2604.00626},
  year={2026}
}

@article{liu2026opsd,
  title={OPSD-V: On-Policy Self-Distillation for Post-Training Few-Step Autoregressive Video Generators},
  author={Liu, Hongyu and Wang, Chun and Gao, Feng and He, Xuanhua and Ma, Yue and Wan, Ziyu and Zhang, Yong and Wei, Xiaoming and Chen, Qifeng},
  journal={arXiv preprint arXiv:2607.08766},
  year={2026}
}

@article{li2026diffusionopd,
  title={DiffusionOPD: A unified perspective of on-policy distillation in diffusion models},
  author={Li, Quanhao and Yu, Junqiu and Jiang, Kaixun and Wei, Yujie and Xing, Zhen and Li, Pandeng and Chu, Ruihang and Zhang, Shiwei and Liu, Yu and Wu, Zuxuan},
  journal={arXiv preprint arXiv:2605.15055},
  year={2026}
}

@article{fang2026flow,
  title={Flow-opd: On-policy distillation for flow matching models},
  author={Fang, Zhen and Huang, Wenxuan and Zeng, Yu and Zhao, Yiming and Chen, Shuang and Feng, Kaituo and Lin, Yunlong and Chen, Lin and Chen, Zehui and Cao, Shaosheng and others},
  journal={arXiv preprint arXiv:2605.08063},
  year={2026}
}

@article{zhou2026danceopd,
  title={DanceOPD: On-Policy Generative Field Distillation},
  author={Zhou, Wei and Zhu, Xiongwei and Xu, Zelin and Dong, Bo and Gong, Lixue and Liang, Yongyuan and Chu, Meng and Qu, Leigang and Kong, Lingdong and Liu, Wei and others},
  journal={arXiv preprint arXiv:2606.27377},
  year={2026}
}

@inproceedings{lin2025stiv,
  title={Stiv: Scalable text and image conditioned video generation},
  author={Lin, Zongyu and Liu, Wei and Chen, Chen and Lu, Jiasen and Hu, Wenze and Fu, Tsu-Jui and Allardice, Jesse and Lai, Zhengfeng and Song, Liangchen and Zhang, Bowen and others},
  booktitle={Proceedings of the IEEE/CVF International Conference on Computer Vision},
  pages={16249--16259},
  year={2025}
}

@article{ma2026scaling,
  title={Scaling Mixture-of-Experts Video Pretraining for Embodied Intelligence},
  author={Ma, Shuailei and Liao, Jiaqi and Wang, Xinyang and Wang, Jingjing and Feng, Chaoran and Hu, Zijing and Bao, Chong and Xi, Zichen and Gan, Yuqi and Wang, Weisen and others},
  journal={arXiv preprint arXiv:2607.07675},
  year={2026}
}

@article{shenfeld2026self,
  title={Self-Distillation Enables Continual Learning},
  author={Shenfeld, Idan and Damani, Mehul and H{\"u}botter, Jonas and Agrawal, Pulkit},
  journal={arXiv preprint arXiv:2601.19897},
  year={2026}
}

@article{yang2026self,
  title={Self-distilled rlvr},
  author={Yang, Chenxu and Qin, Chuanyu and Si, Qingyi and Chen, Minghui and Gu, Naibin and Yao, Dingyu and Lin, Zheng and Wang, Weiping and Wang, Jiaqi and Duan, Nan},
  journal={arXiv preprint arXiv:2604.03128},
  year={2026}
}

@article{zhao2026self,
  title={Self-Distilled Reasoner: On-Policy Self-Distillation for Large Language Models},
  author={Zhao, Siyan and Xie, Zhihui and Liu, Mengchen and Huang, Jing and Pang, Guan and Chen, Feiyu and Grover, Aditya},
  journal={arXiv preprint arXiv:2601.18734},
  year={2026}
}

@inproceedings{he2026self,
  title={Self-distillation zero: Self-revision turns binary rewards into dense supervision},
  author={He, Yinghui and Kaur, Simran and Bhaskar, Adithya and Yang, Yongjin and Liu, Jiarui and Ri, Narutatsu and Fowl, Liam H and Panigrahi, Abhishek and Chen, Danqi and Arora, Sanjeev},
  booktitle={ICML 2026 Workshop on Foundations of Deep Generative Models: Understanding Memorization, Generalization, and Reasoning},
  year={2026}
}

@article{jiang2025no,
  title={No other representation component is needed: Diffusion transformers can provide representation guidance by themselves},
  author={Jiang, Dengyang and Wang, Mengmeng and Li, Liuzhuozheng and Zhang, Lei and Wang, Haoyu and Wei, Wei and Dai, Guang and Zhang, Yanning and Wang, Jingdong},
  journal={arXiv preprint arXiv:2505.02831},
  year={2025}
}

@article{wang2025promptenhancer,
  title={Promptenhancer: A simple approach to enhance text-to-image models via chain-of-thought prompt rewriting},
  author={Wang, Linqing and Xing, Ximing and Cheng, Yiji and Zhao, Zhiyuan and Li, Donghao and Hang, Tiankai and Tao, Jiale and Wang, Qixun and Li, Ruihuang and Chen, Comi and others},
  journal={arXiv preprint arXiv:2509.04545},
  year={2025}
}

@inproceedings{cheng2025vpo,
  title={Vpo: Aligning text-to-video generation models with prompt optimization},
  author={Cheng, Jiale and Lyu, Ruiliang and Gu, Xiaotao and Liu, Xiao and Xu, Jiazheng and Lu, Yida and Teng, Jiayan and Yang, Zhuoyi and Dong, Yuxiao and Tang, Jie and others},
  booktitle={Proceedings of the IEEE/CVF International Conference on Computer Vision},
  pages={15636--15645},
  year={2025}
}

@inproceedings{meng2023distillation,
  title={On distillation of guided diffusion models},
  author={Meng, Chenlin and Rombach, Robin and Gao, Ruiqi and Kingma, Diederik and Ermon, Stefano and Ho, Jonathan and Salimans, Tim},
  booktitle={Proceedings of the IEEE/CVF conference on computer vision and pattern recognition},
  pages={14297--14306},
  year={2023}
}

@article{salimans2022progressive,
  title={Progressive distillation for fast sampling of diffusion models},
  author={Salimans, Tim and Ho, Jonathan},
  journal={arXiv preprint arXiv:2202.00512},
  year={2022}
}

@article{flux1-lite,
  title={Flux.1 Lite: Distilling Flux1.dev for Efficient Text-to-Image Generation},
  author={Daniel Verdú, Javier Martín},
  email={dverdu@freepik.com, javier.martin@freepik.com},
  year={2024},
}

@article{yu2024representation,
  title={Representation alignment for generation: Training diffusion transformers is easier than you think},
  author={Yu, Sihyun and Kwak, Sangkyung and Jang, Huiwon and Jeong, Jongheon and Huang, Jonathan and Shin, Jinwoo and Xie, Saining},
  journal={arXiv preprint arXiv:2410.06940},
  year={2024}
}

@inproceedings{fang2025tinyfusion,
  title={Tinyfusion: Diffusion transformers learned shallow},
  author={Fang, Gongfan and Li, Kunjun and Ma, Xinyin and Wang, Xinchao},
  booktitle={Proceedings of the Computer Vision and Pattern Recognition Conference},
  pages={18144--18154},
  year={2025}
}

@article{wu2026representation,
  title={Representation entanglement for generation: Training diffusion transformers is much easier than you think},
  author={Wu, Ge and Zhang, Shen and Shi, Ruijing and Gao, Shanghua and Chen, Zhenyuan and Wang, Lei and Chen, Zhaowei and Gao, Hongcheng and Tang, Yao and Cheng, Ming-Ming and others},
  journal={Advances in Neural Information Processing Systems},
  volume={38},
  pages={7714--7743},
  year={2026}
}

@article{lu2025onpolicydistillation,
  author = {Kevin Lu and Thinking Machines Lab},
  title = {On-Policy Distillation},
  journal = {Thinking Machines Lab: Connectionism},
  year = {2025},
  note = {https://thinkingmachines.ai/blog/on-policy-distillation},
  doi = {10.64434/tml.20251026},
}

@inproceedings{jang2026stable,
  title={Stable on-policy distillation through adaptive target reformulation},
  author={Jang, Ijun and Yeom, Jewon and Yeo, Juan and Lim, Hyunggyu and Kim, Taesup},
  booktitle={Findings of the Association for Computational Linguistics: ACL 2026},
  pages={42217--42227},
  year={2026}
}

@misc{flux-2-2025,
    author={Black Forest Labs},
    title={{FLUX.2: Frontier Visual Intelligence}},
    year={2025},
    howpublished={\url{https://bfl.ai/blog/flux-2}},
}

@article{wu2025hunyuanvideo,
  title={Hunyuanvideo 1.5 technical report},
  author={Wu, Bing and Zou, Chang and Li, Changlin and Huang, Duojun and Yang, Fang and Tan, Hao and Peng, Jack and Wu, Jianbing and Xiong, Jiangfeng and Jiang, Jie and others},
  journal={arXiv preprint arXiv:2511.18870},
  year={2025}
}

@misc{qwen36_27b,
    title = {{Qwen3.6-27B}: Flagship-Level Coding in a 27B Dense Model},
    url = {https://qwen.ai/blog?id=qwen3.6-27b},
    author = {{Qwen Team}},
    month = {April},
    year = {2026}
}

\newpage
\appendix

\section{Appendix}
In the appendix, we provide 
additional implementation details (Section~\ref{sec:details}), 
additional qualitative samples (Section~\ref{sec:add-qualitative}), 
additional experimental results (Section~\ref{sec:add-quantitative}), 
all text prompts used in both the main paper and appendix (Section~\ref{sec:prompt}), 
examples of privileged conditions (Section~\ref{sec:condition}), 
the ethical statement (Section~\ref{sec:ethical}), 
the reproducibility statement (Section~\ref{sec:reproducibility}),  
as well as the declaration on LLM usage (Section~\ref{sec:declaration}), 
as a supplement to the main paper. 

\section{Additional Implementation Details}\label{sec:details}

\subsection{Hyperparameter Configuration}

The detailed hyperparameter settings used in this paper are listed in Tab.~\ref{tab:hyperparams}.
Unless otherwise specified, these parameters remain consistent across all experiments.

\begin{table}[ht]
\caption{Hyperparameter settings in our experiments.}
\centering
\resizebox{0.9\linewidth}{!}{ 
\begin{tabular}{lclc}
\toprule
\textbf{Parameter} & \textbf{Value} & \textbf{Parameter} & \textbf{Value} \\
\midrule
Base models & WAN-2.2, LTX-2.3 & LoRA settings & $r=32, \alpha=64$ \\
Training frames & $33$ & Anchor-trajectory length (WAN-2.2) & $50$ \\
Train batch size (per GPU) & $1$ & Anchor-trajectory length (LTX-2.3) & $30$ \\
The number of GPUs & $8$ & Optimizer & AdamW \\
Training steps & $500$ & Learning rate & $1\times 10^{-4}$ \\
Training guidance scale & $1.0$ & Weight decay & $0.0$ \\
Anchor steps ($|\mathcal{A}|$) & 6 & Mixed precision & \texttt{bfloat16} \\
Resolution (WAN-2.2) & $1280\times704$ & EMA decay rate & $0.999$ \\
Resolution (LTX-2.3) & $768\times512$ & Sub-trajectory length ($K$) & $3$ \\
\bottomrule
\label{tab:hyperparams}
\end{tabular}
}
\end{table}

\subsection{Training Convergence}
\label{sec:convergence}

\begin{wrapfigure}{r}{0.42\linewidth}
    \vspace{-3.5em}
    \centering
    \includegraphics[width=\linewidth]{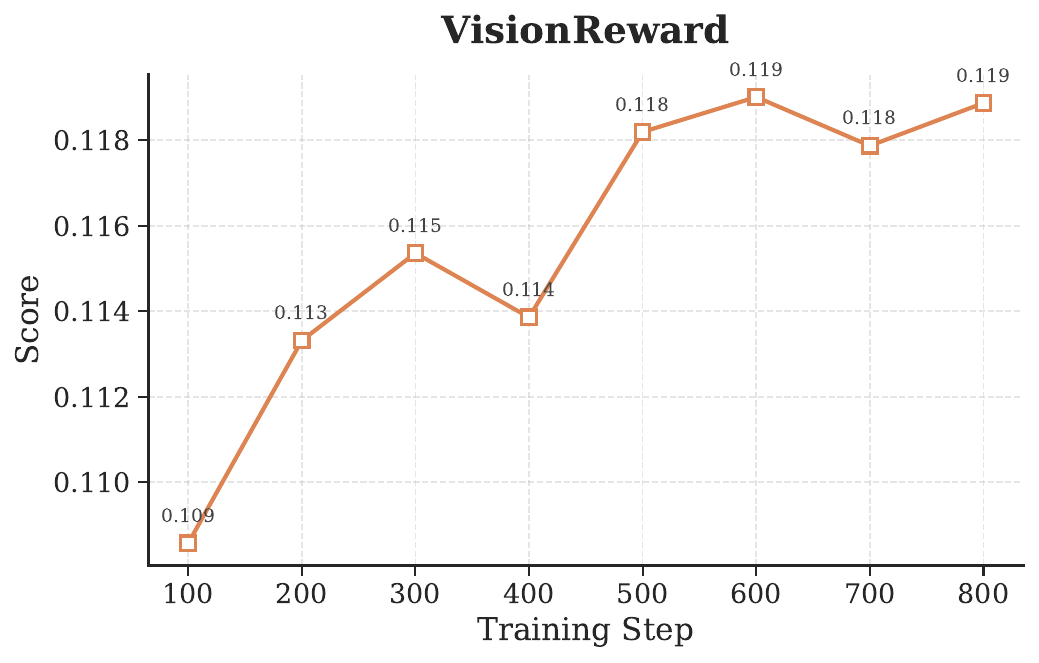}
    \vspace{-2em}
    \caption{
        \textbf{Training curve of HPSD.}
    }
    \label{fig:convergence}
\end{wrapfigure}

To determine the training budget, we investigate the convergence behavior of HPSD by monitoring its VisionReward score on WAN-2.2. 
As illustrated in Fig.~\ref{fig:convergence}, the metric improves rapidly in the early stage and plateaus around 500 steps, with negligible fluctuation thereafter, indicating that the model has reached an approximately converged state. 
Based on this empirical observation, we adopt 500 steps (trained on 8 GPUs) as the default configuration for all our main experiments.

\subsection{Prompts for LLM Rewriter}
\label{sec:llm_prompts}

\begin{figure}[t]
    \centering
    \includegraphics[width=0.75\linewidth]{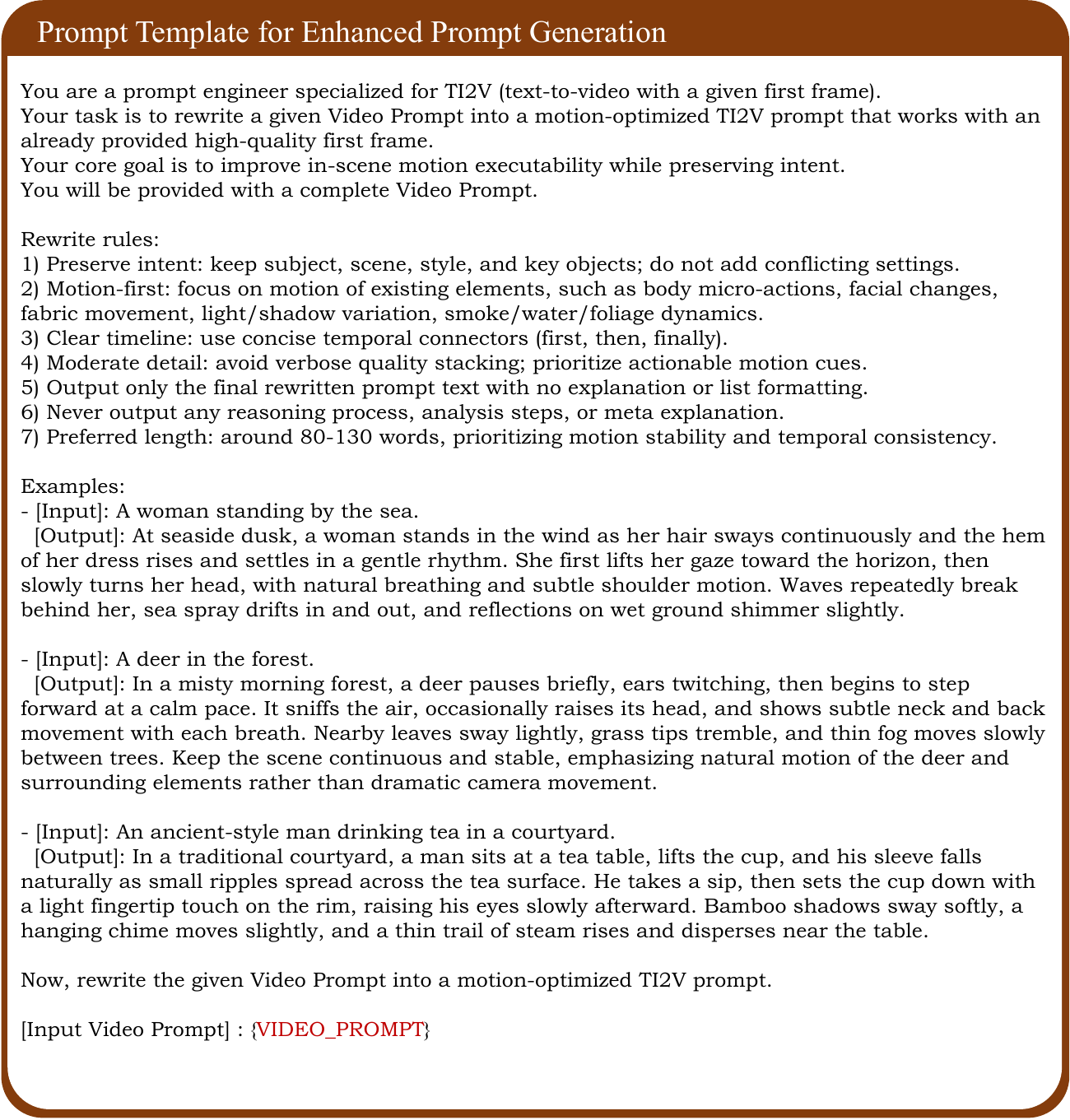}
    \caption{
        \textbf{Prompt Template for Enhanced Prompt Generation.} 
        }
    \label{fig:enhanced-prompt-template}
\end{figure}

\begin{figure}[t]
    \centering
    \includegraphics[width=0.75\linewidth]{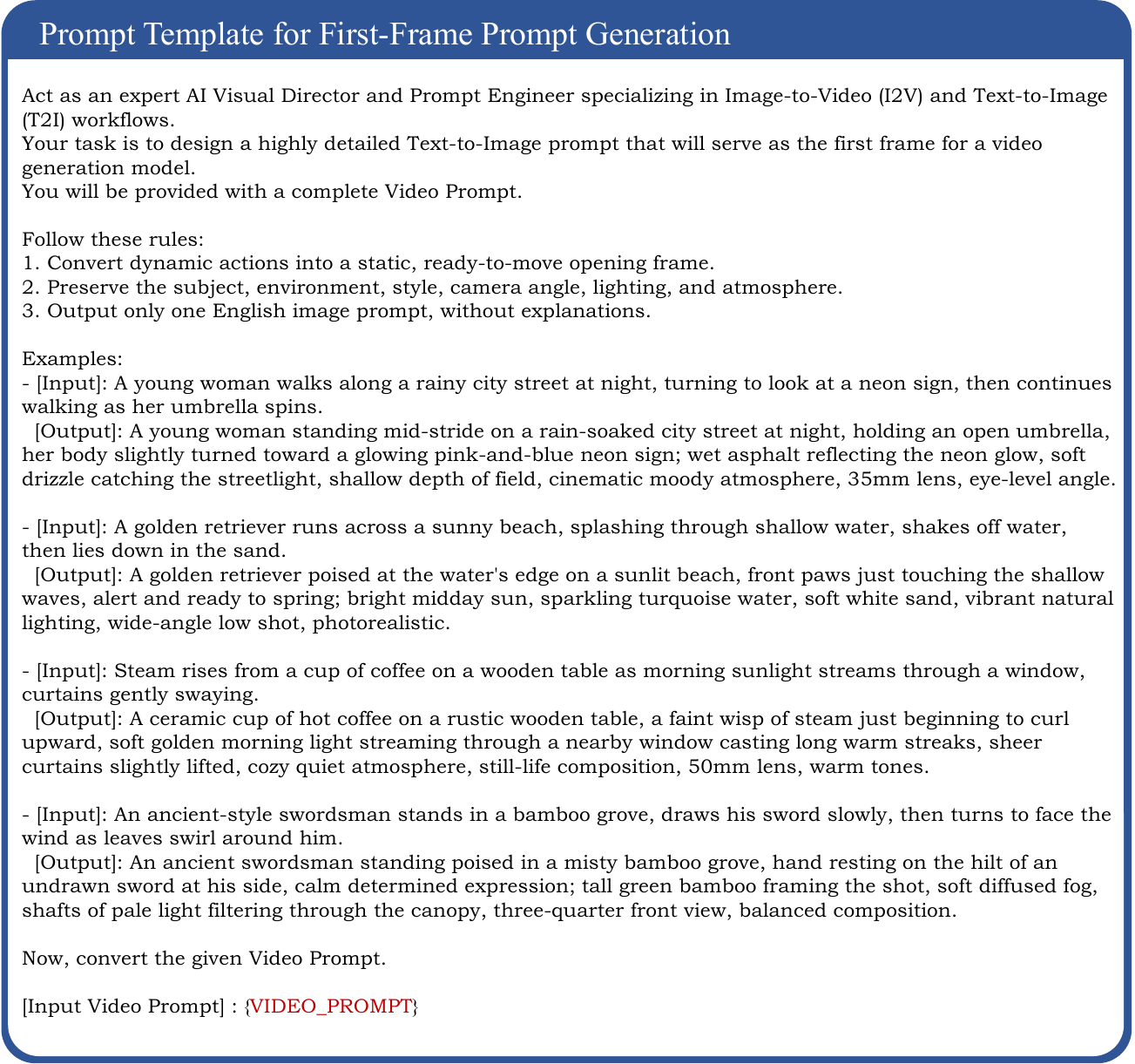}
    \caption{
        \textbf{Prompt Template for First-Frame Prompt Generation.} 
        }
    \label{fig:ff-prompt-template}
\end{figure}

To construct the privileged conditions, we employ an external LLM (Qwen3.6-27B) as the prompt rewriter. The specific system prompts designed for these tasks are provided in Fig.~\ref{fig:enhanced-prompt-template} and Fig.~\ref{fig:ff-prompt-template}.

\noindent\textbf{Enhanced Prompt Generation}.
In the full HPSD pipeline, the enhanced prompt $c_\text{txt}^{+}$ is utilized alongside the privileged first-frame condition $c_\text{img}$. Since the visual aesthetics and spatial layout of the video are already anchored by this high-quality first frame, we design a system prompt that directs the LLM to focus primarily on temporal dynamics, as illustrated in Fig.~\ref{fig:enhanced-prompt-template}. 

\noindent\textbf{First-Frame Prompt Generation}.
To guide the auxiliary T2I model in synthesizing a structurally sound and aesthetically pleasing first frame $c_\text{img}$, the original video prompt $c_\text{txt}$ is converted into an image generation prompt $c_\text{ff}$. 
Specifically, the LLM is instructed to translate dynamic actions into a static, ready-to-move opening posture, as detailed in the prompt template of Fig.~\ref{fig:ff-prompt-template}.

\clearpage
\newpage

\section{Additional Qualitative Samples}\label{sec:add-qualitative}

\begin{figure}[t]
    \centering
    \includegraphics[width=0.95\linewidth]{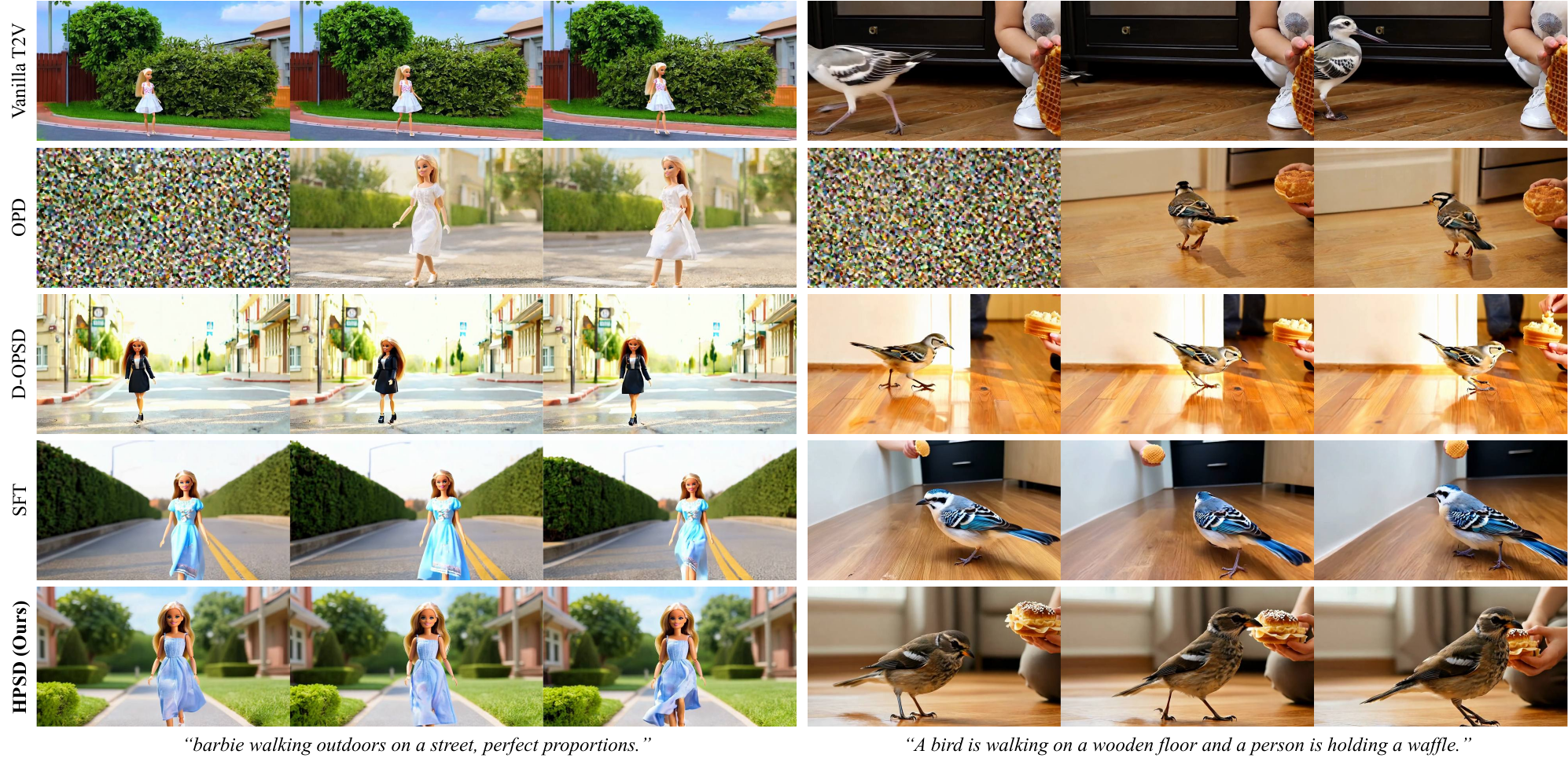}
    \vspace{-1em}
    \caption{
        \textbf{More qualitative comparison results on WAN-2.2 (1/6).} 
        Best viewed zoomed in.
        }
    \label{fig:wan22-add-1}
\end{figure}

\begin{figure}[t]
    \centering
    \includegraphics[width=0.95\linewidth]{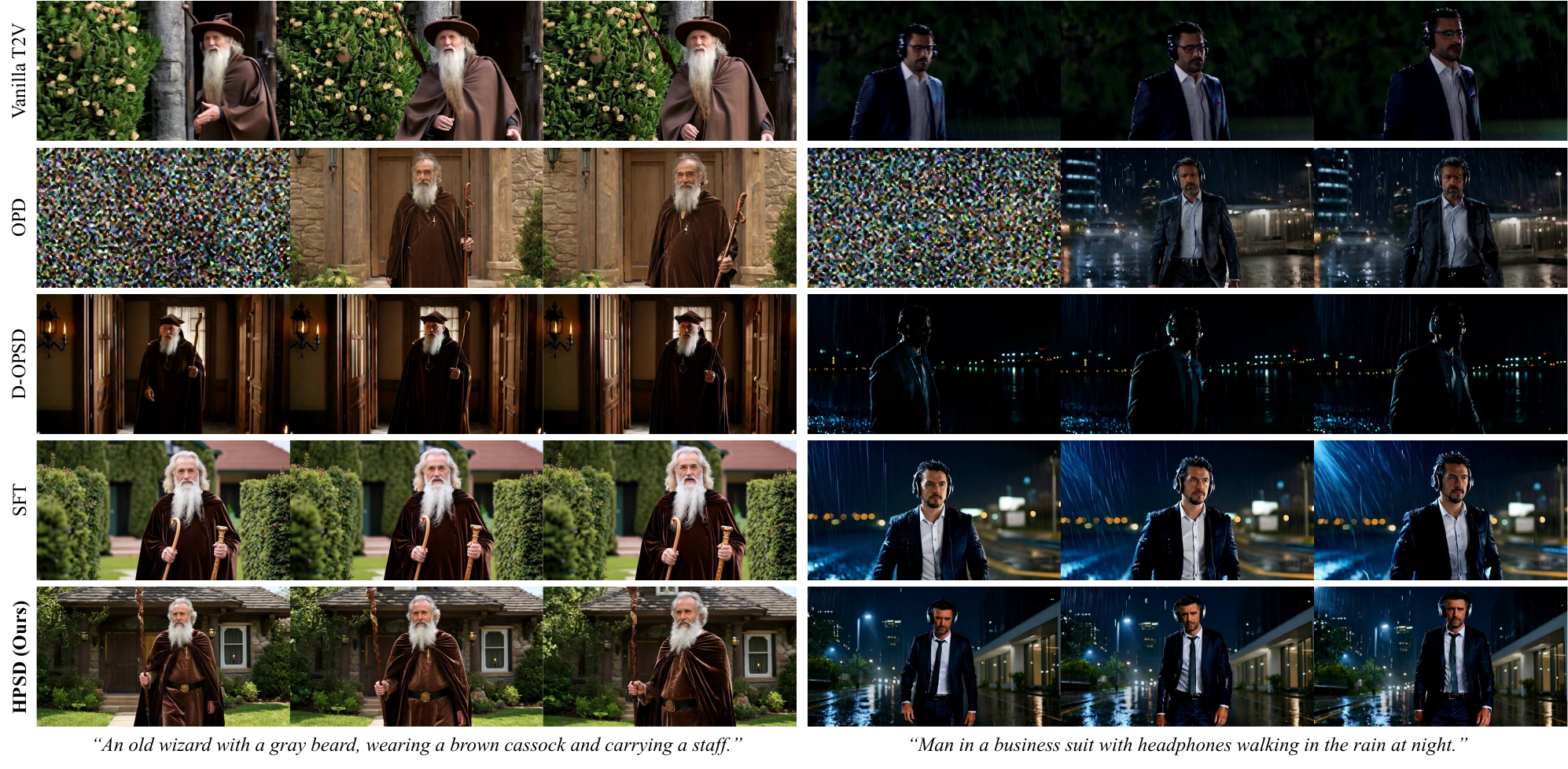}
    \vspace{-1em}
    \caption{
        \textbf{More qualitative comparison results on WAN-2.2 (2/6).} 
        Best viewed zoomed in.
        }
    \label{fig:wan22-add-2}
\end{figure}

\begin{figure}[t]
    \centering
    \includegraphics[width=0.95\linewidth]{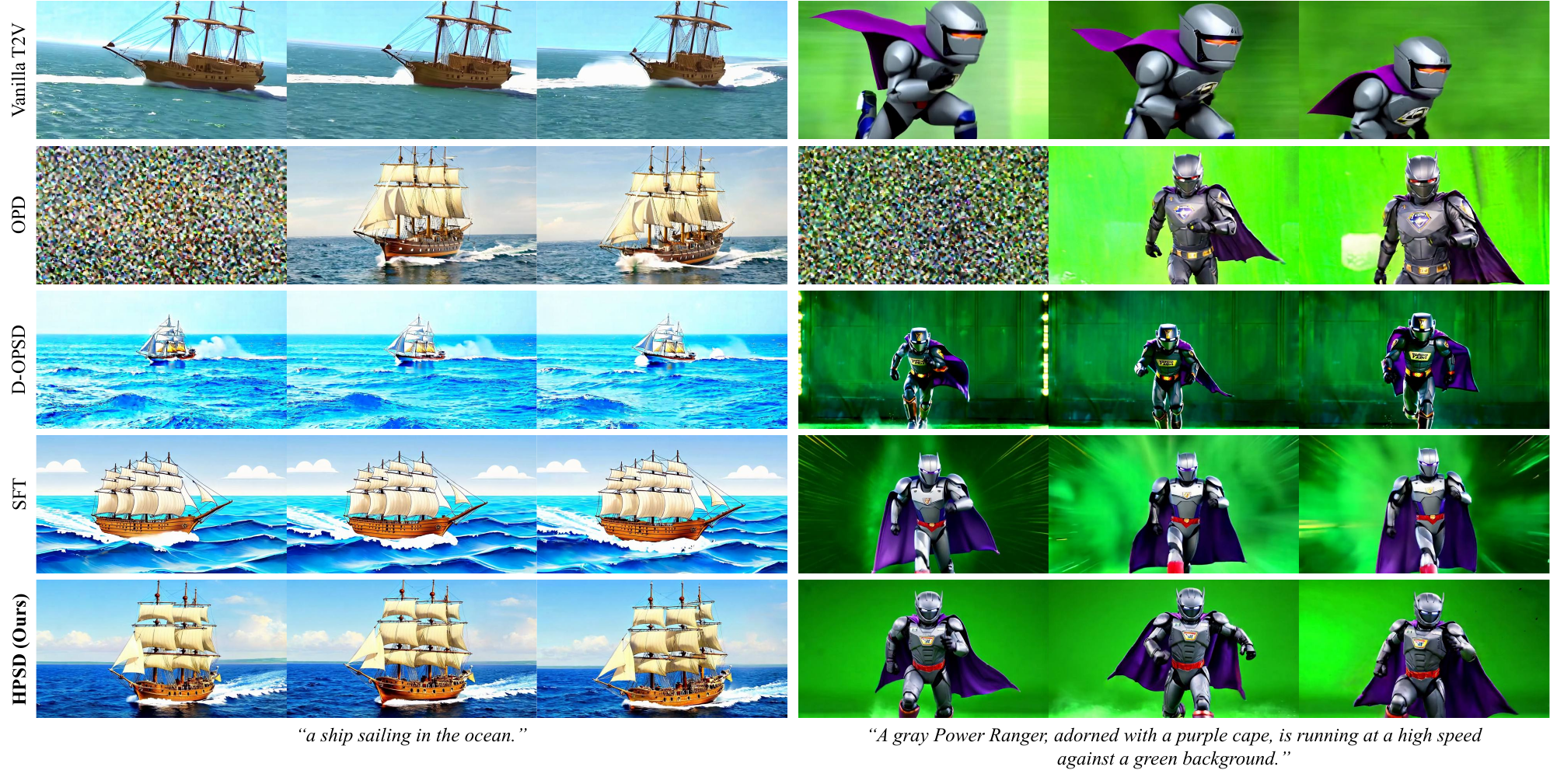}
    \vspace{-1em}
    \caption{
        \textbf{More qualitative comparison results on WAN-2.2 (3/6).} 
        Best viewed zoomed in.
        }
    \label{fig:wan22-add-3}
\end{figure}

\begin{figure}[t]
    \centering
    \includegraphics[width=0.95\linewidth]{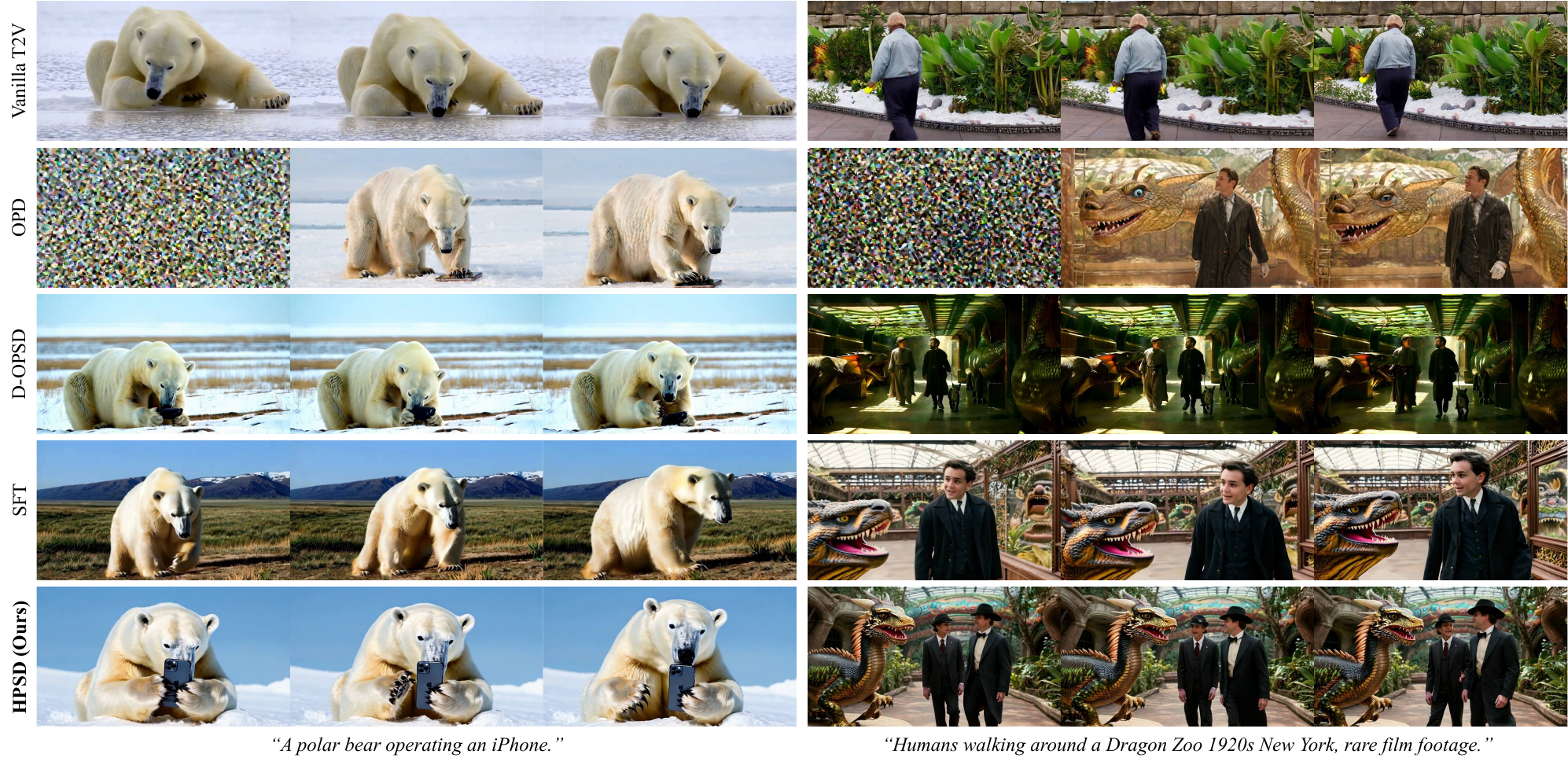}
    \vspace{-1em}
    \caption{
        \textbf{More qualitative comparison results on WAN-2.2 (4/6).} 
        Best viewed zoomed in.
        }
    \label{fig:wan22-add-4}
\end{figure}

\begin{figure}[t]
    \centering
    \includegraphics[width=0.95\linewidth]{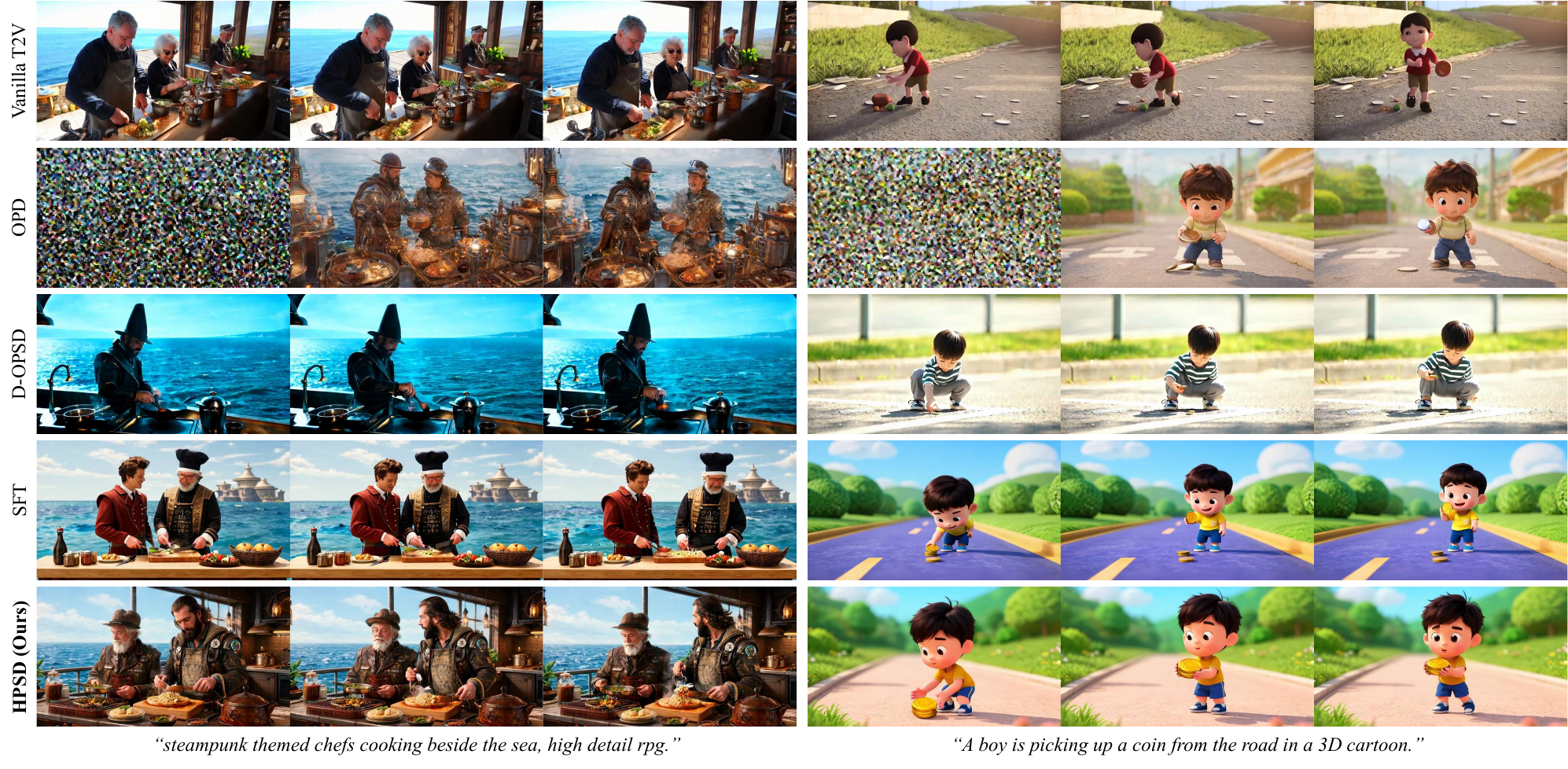}
    \vspace{-1em}
    \caption{
        \textbf{More qualitative comparison results on WAN-2.2 (5/6).} 
        Best viewed zoomed in.
        }
    \label{fig:wan22-add-5}
\end{figure}

\begin{figure}[t]
    \centering
    \includegraphics[width=0.95\linewidth]{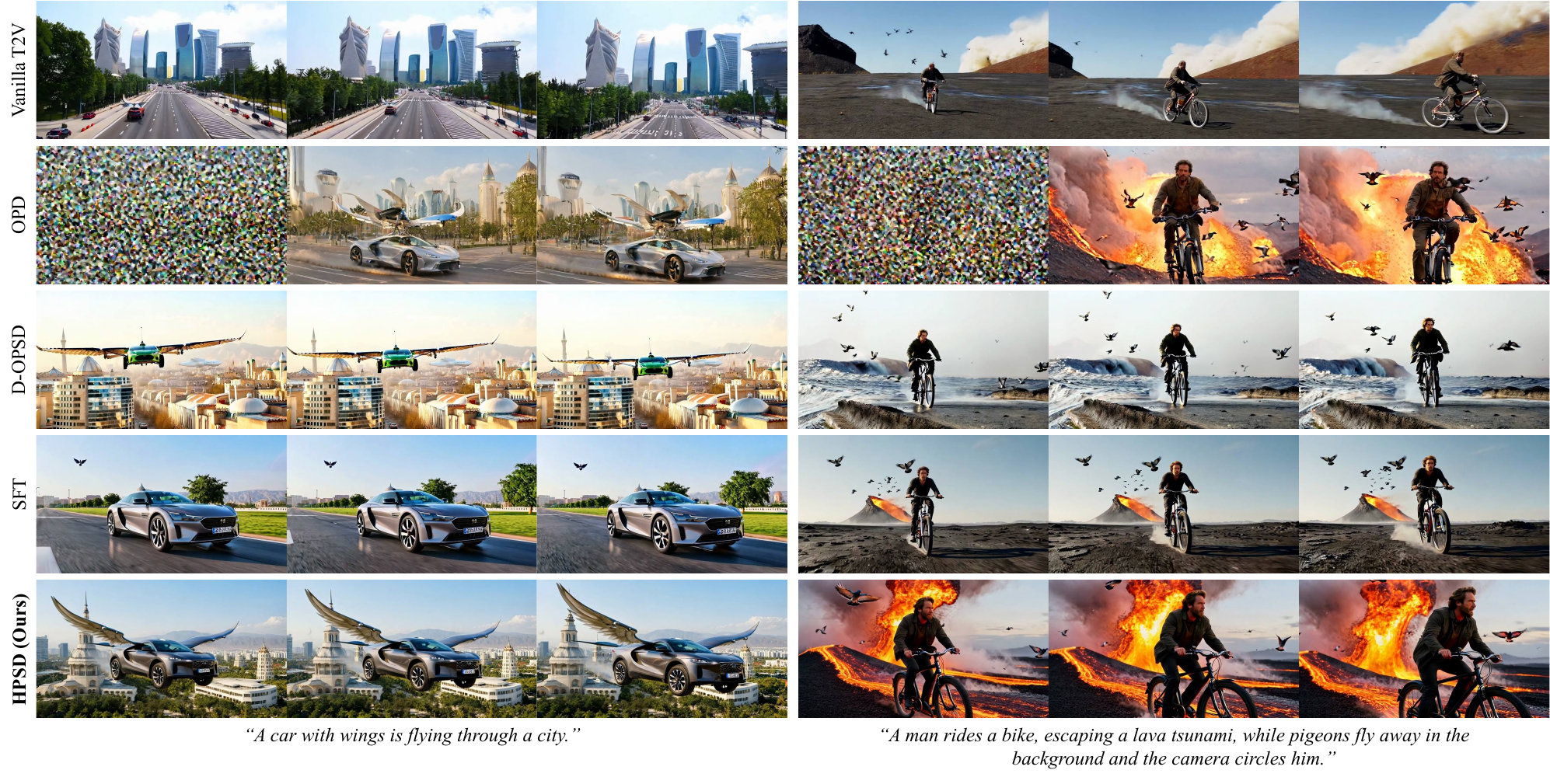}
    \vspace{-1em}
    \caption{
        \textbf{More qualitative comparison results on WAN-2.2 (6/6).} 
        Best viewed zoomed in.
        }
    \label{fig:wan22-add-6}
\end{figure}

\begin{figure}[t]
    \centering
    \includegraphics[width=1.0\linewidth]{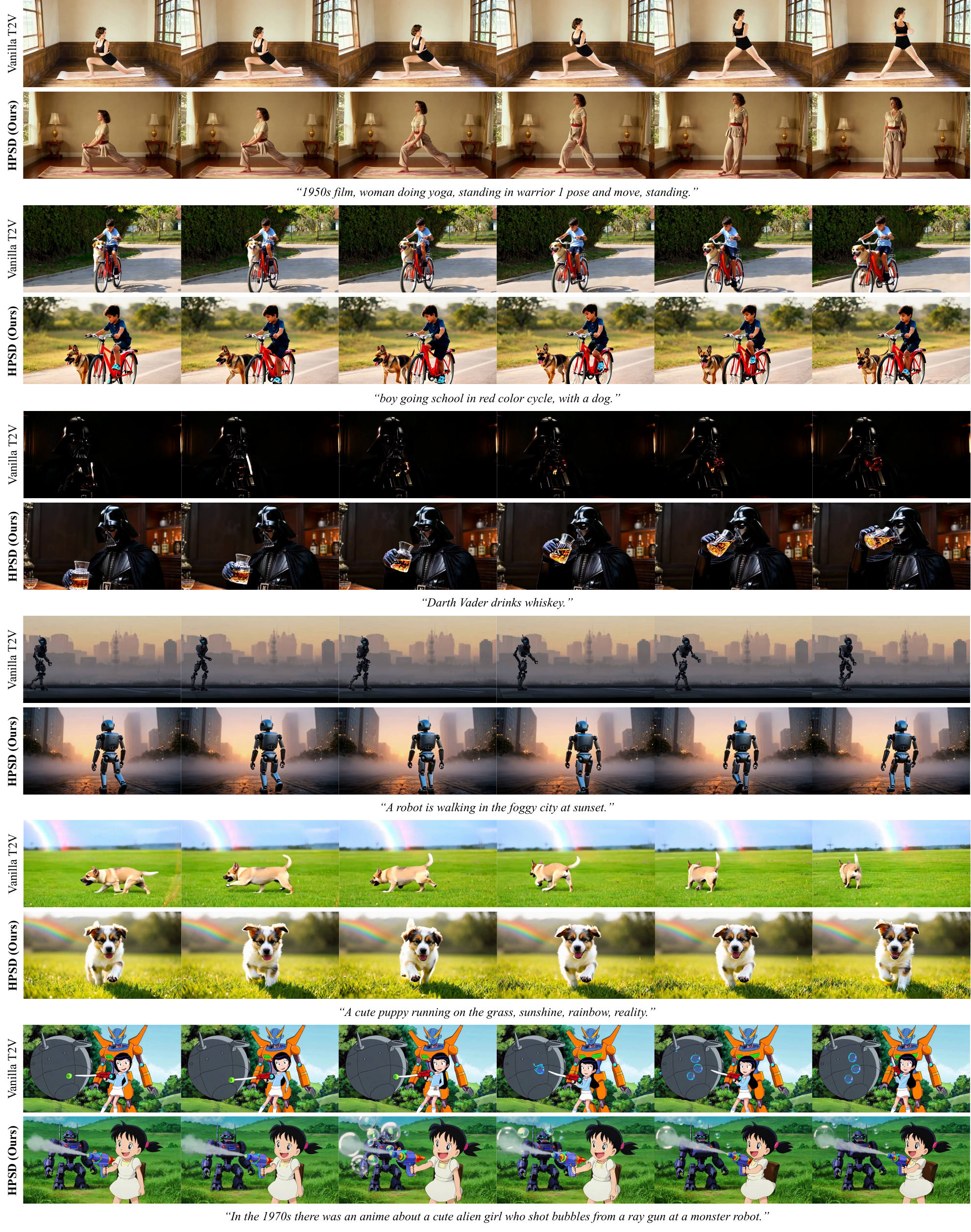}
    \vspace{-1em}
    \caption{
        \textbf{Comparison between HPSD and vanilla WAN-2.2 (1/3).} 
        Best viewed zoomed in.
        }
    \label{fig:wan22-two-1}
\end{figure}

\begin{figure}[t]
    \centering
    \includegraphics[width=1.0\linewidth]{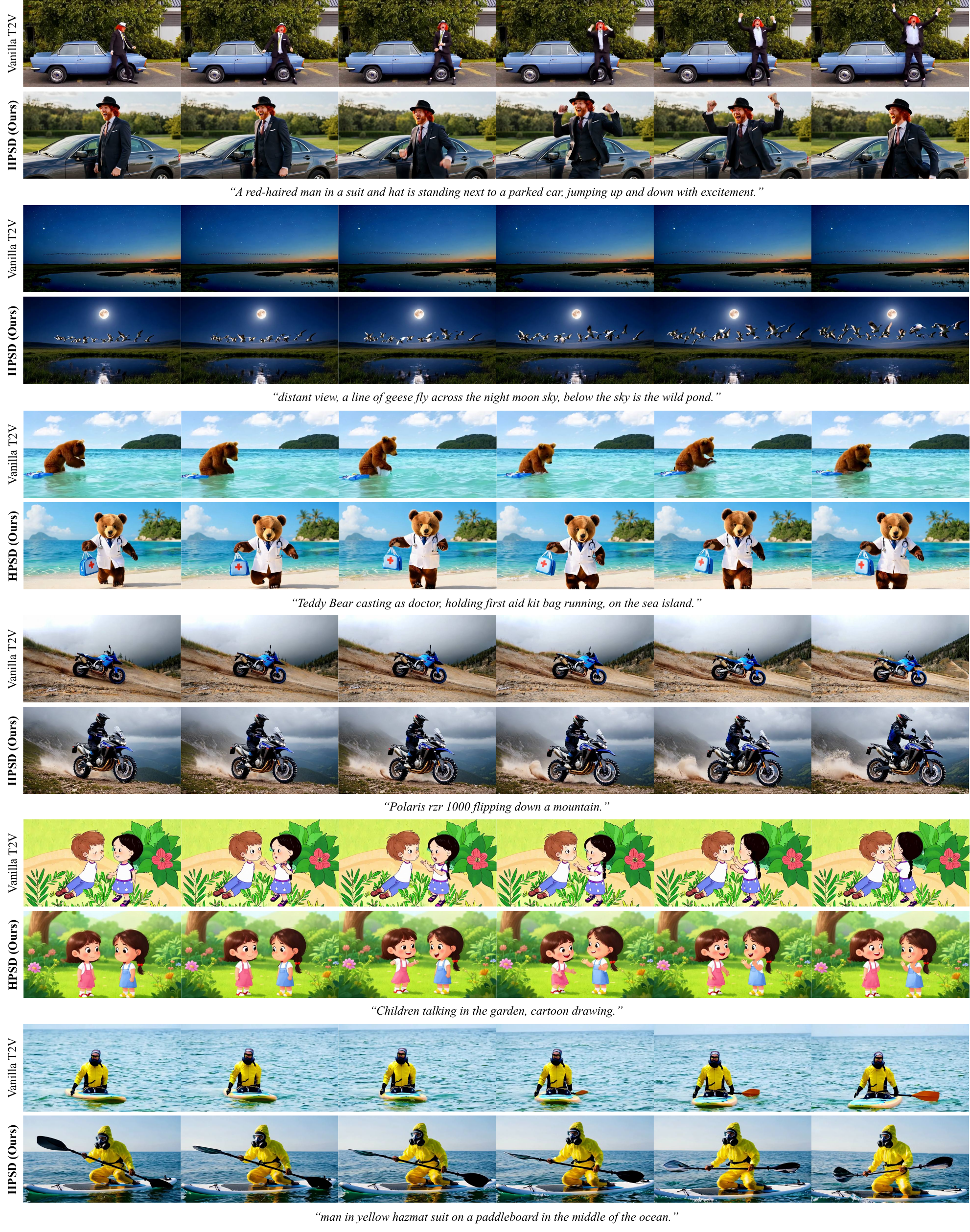}
    \vspace{-1em}
    \caption{
        \textbf{Comparison between HPSD and vanilla WAN-2.2 (2/3).} 
        Best viewed zoomed in.
        }
    \label{fig:wan22-two-2}
\end{figure}

\begin{figure}[t]
    \centering
    \includegraphics[width=1.0\linewidth]{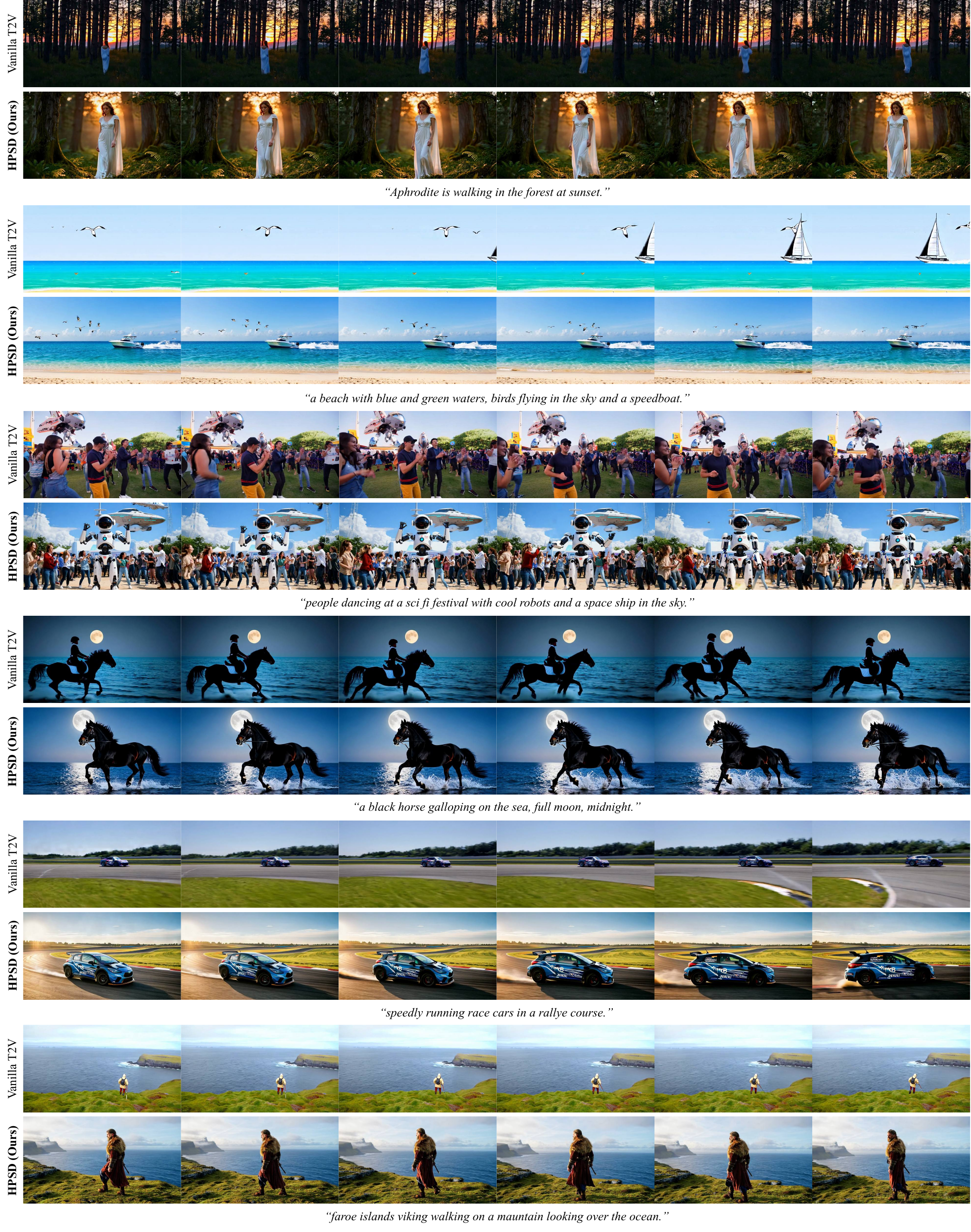}
    \vspace{-1em}
    \caption{
        \textbf{Comparison between HPSD and vanilla WAN-2.2 (3/3).} 
        Best viewed zoomed in.
        }
    \label{fig:wan22-two-3}
\end{figure}

\begin{figure}[t]
    \centering
    \includegraphics[width=0.8\linewidth]{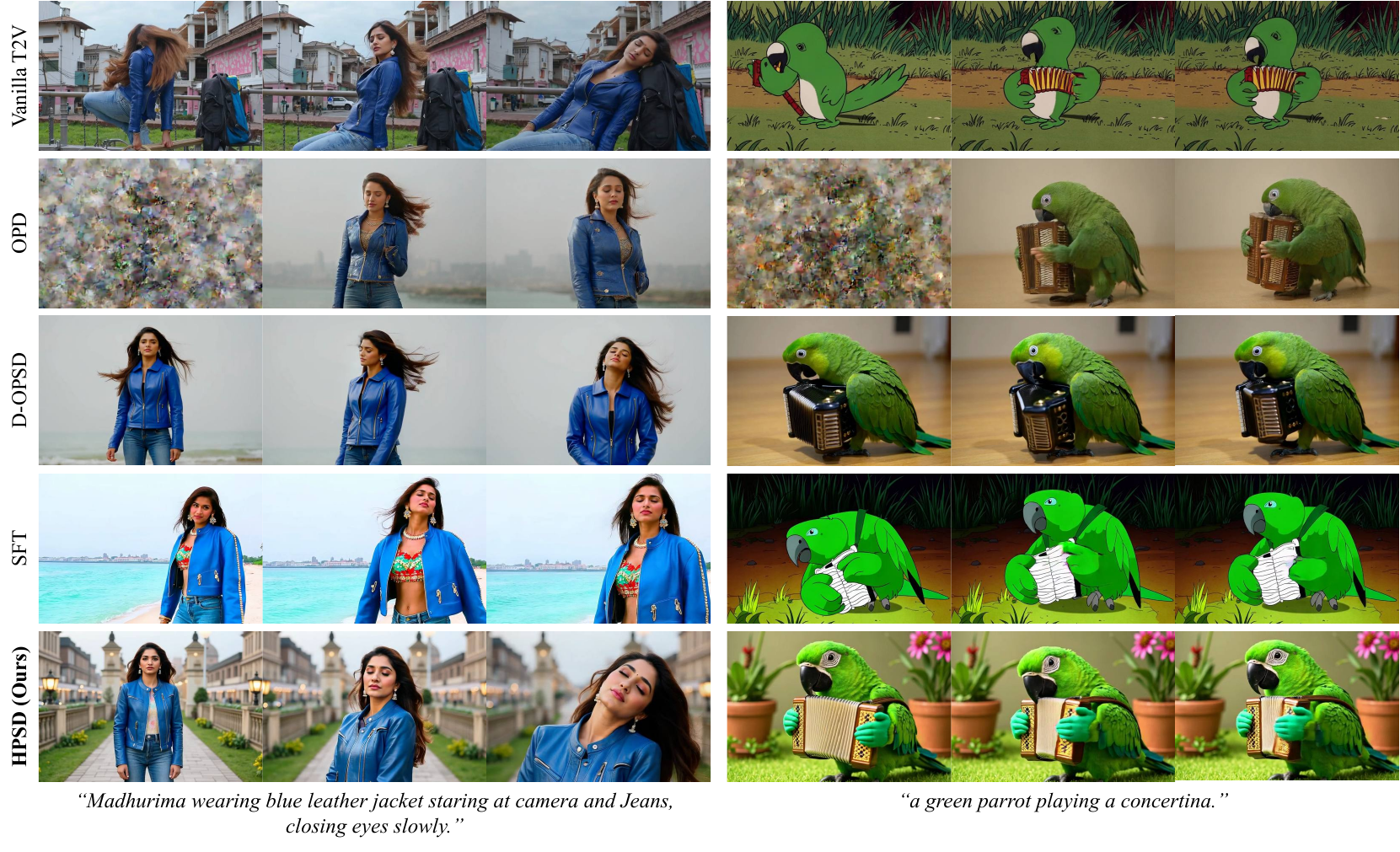}
    \vspace{-1em}
    \caption{
        \textbf{More qualitative comparison results on LTX-2.3 (1/6).} 
        Best viewed zoomed in.
        }
    \label{fig:ltx23-add-1}
\end{figure}

\begin{figure}[t]
    \centering
    \includegraphics[width=0.8\linewidth]{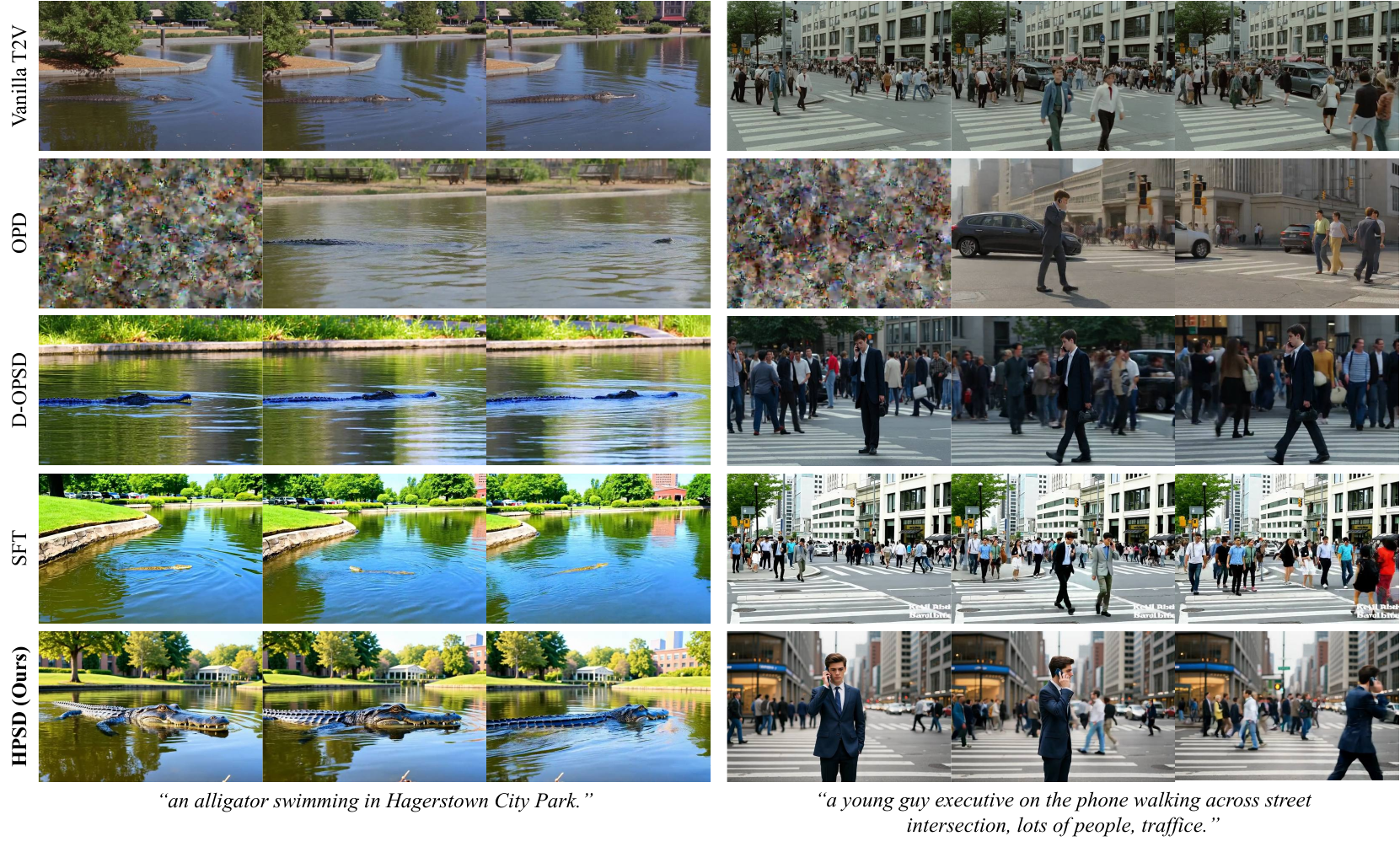}
    \vspace{-1em}
    \caption{
        \textbf{More qualitative comparison results on LTX-2.3 (2/6).} 
        Best viewed zoomed in.
        }
    \label{fig:ltx23-add-2}
\end{figure}

\begin{figure}[t]
    \centering
    \includegraphics[width=0.8\linewidth]{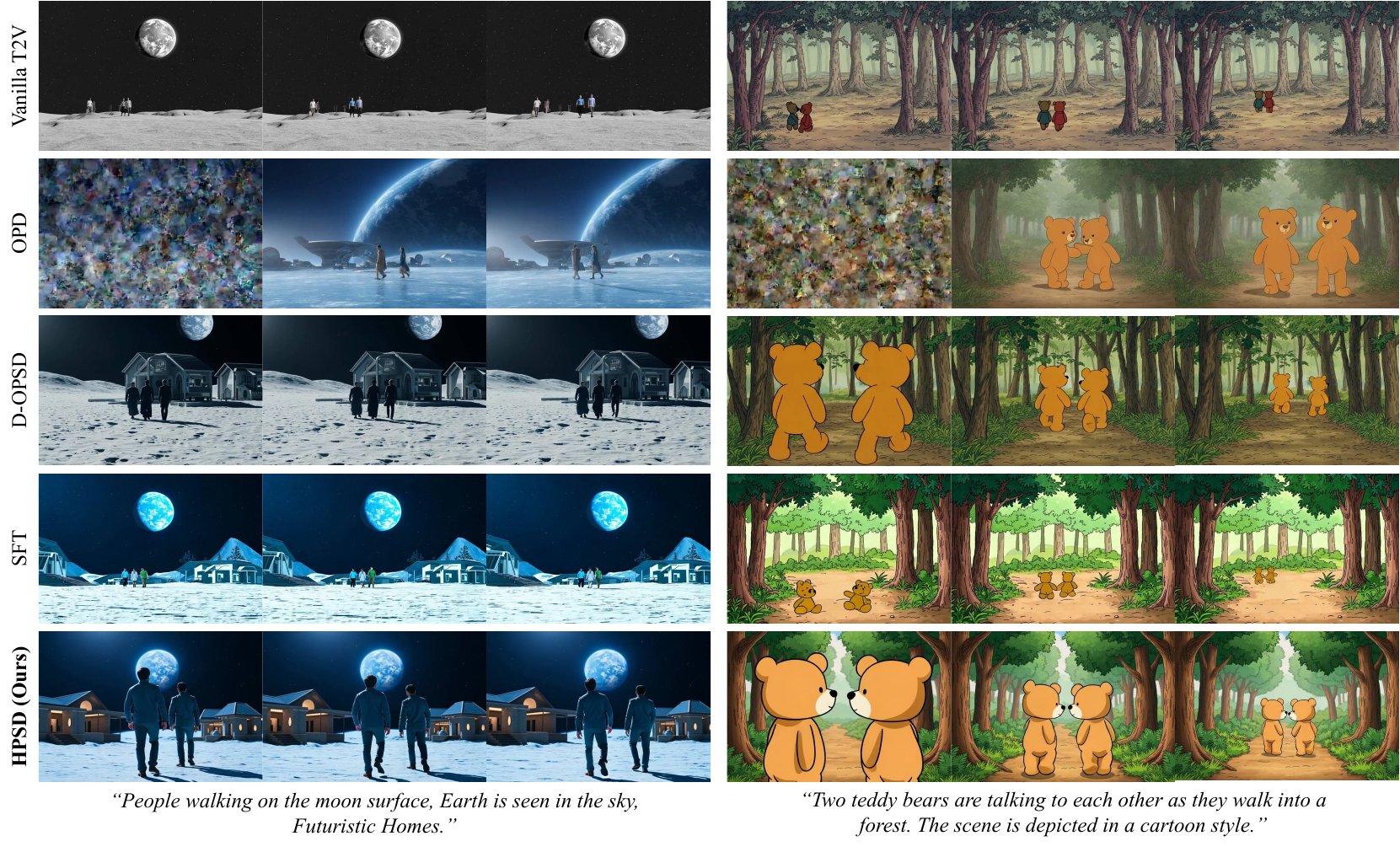}
    \vspace{-1em}
    \caption{
        \textbf{More qualitative comparison results on LTX-2.3 (3/6).} 
        Best viewed zoomed in.
        }
    \label{fig:ltx23-add-3}
\end{figure}

\begin{figure}[t]
    \centering
    \includegraphics[width=0.8\linewidth]{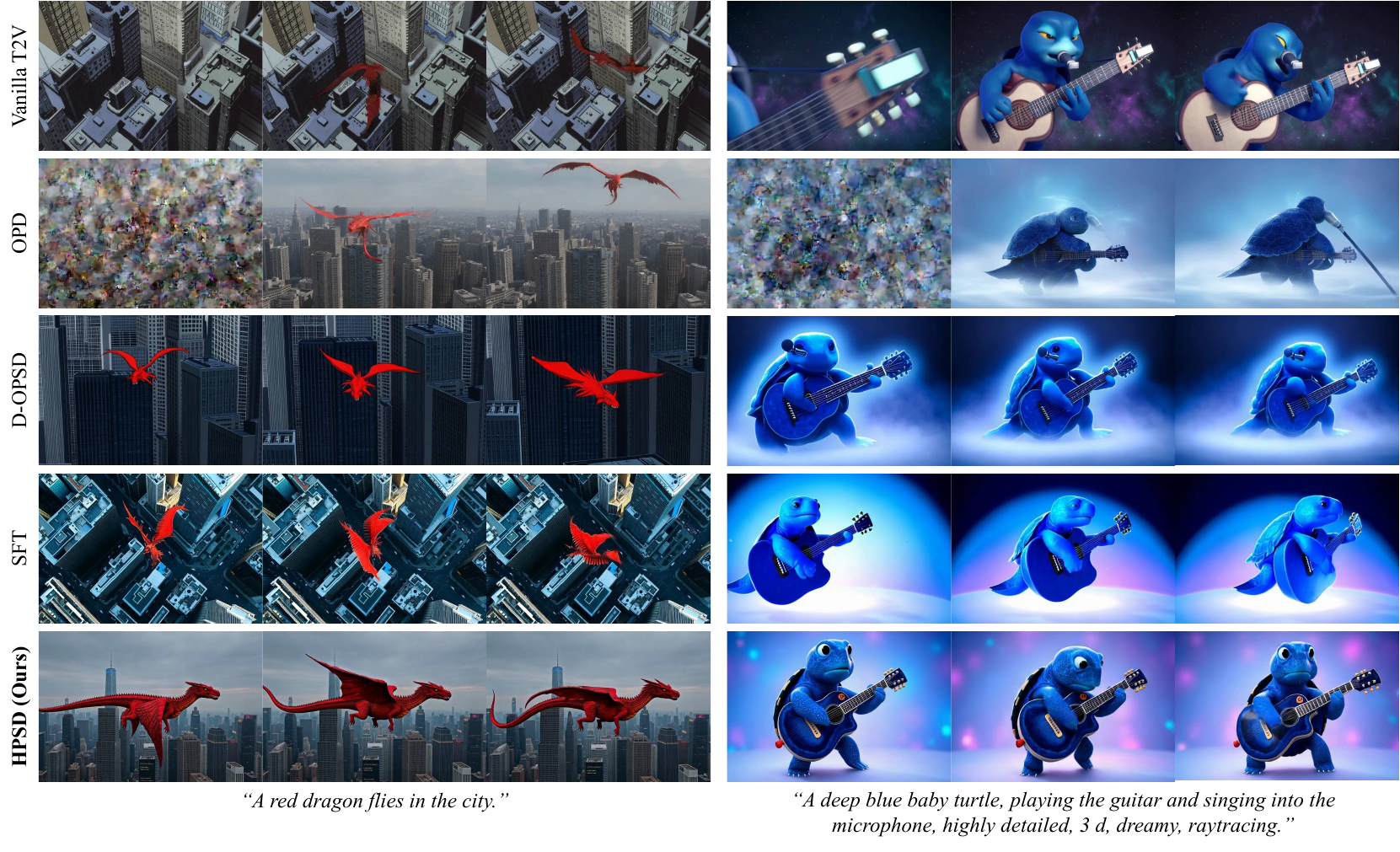}
    \vspace{-1em}
    \caption{
        \textbf{More qualitative comparison results on LTX-2.3 (4/6).} 
        Best viewed zoomed in.
        }
    \label{fig:ltx23-add-4}
\end{figure}

\begin{figure}[t]
    \centering
    \includegraphics[width=0.8\linewidth]{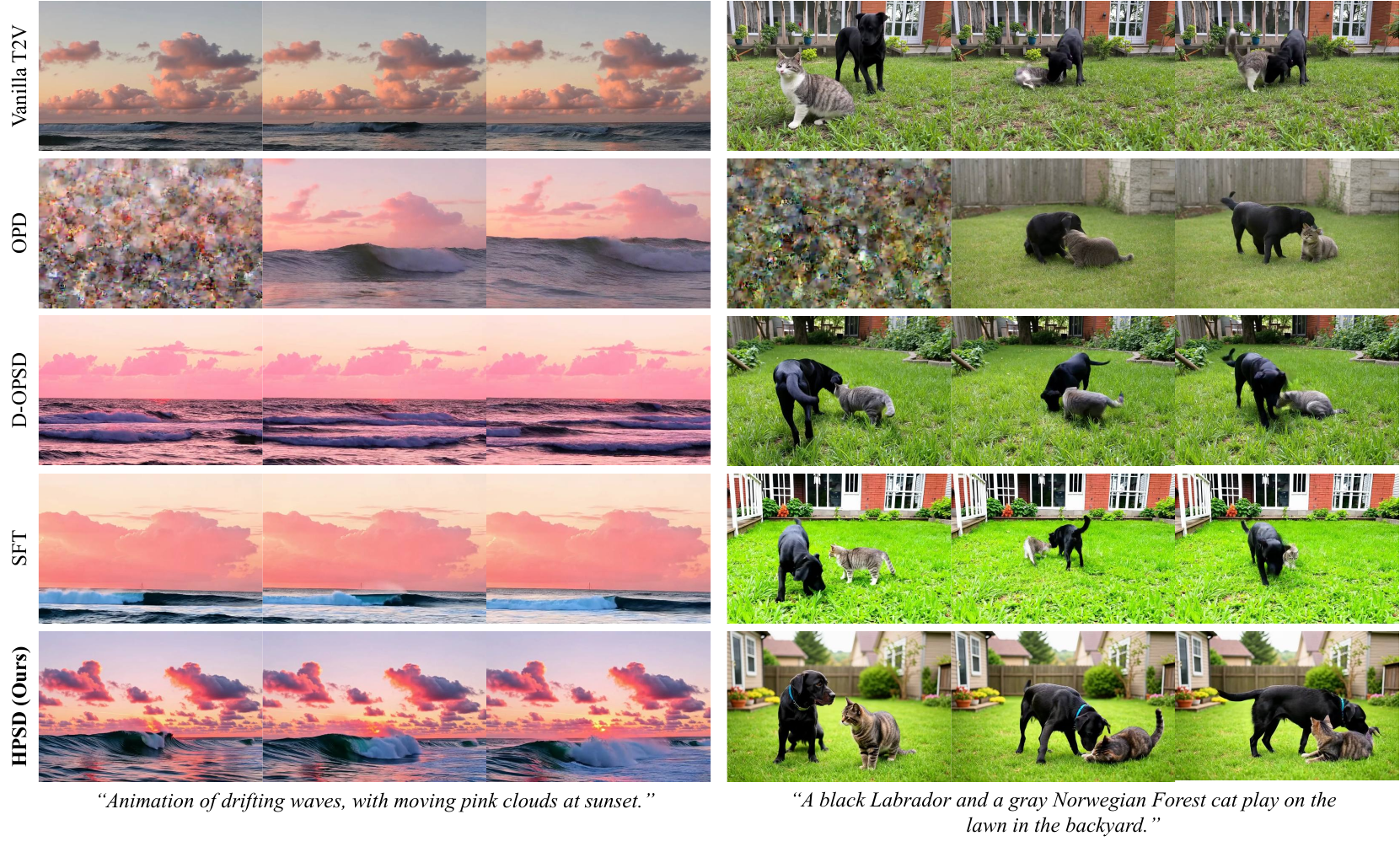}
    \vspace{-1em}
    \caption{
        \textbf{More qualitative comparison results on LTX-2.3 (5/6).} 
        Best viewed zoomed in.
        }
    \label{fig:ltx23-add-5}
\end{figure}

\begin{figure}[t]
    \centering
    \includegraphics[width=0.8\linewidth]{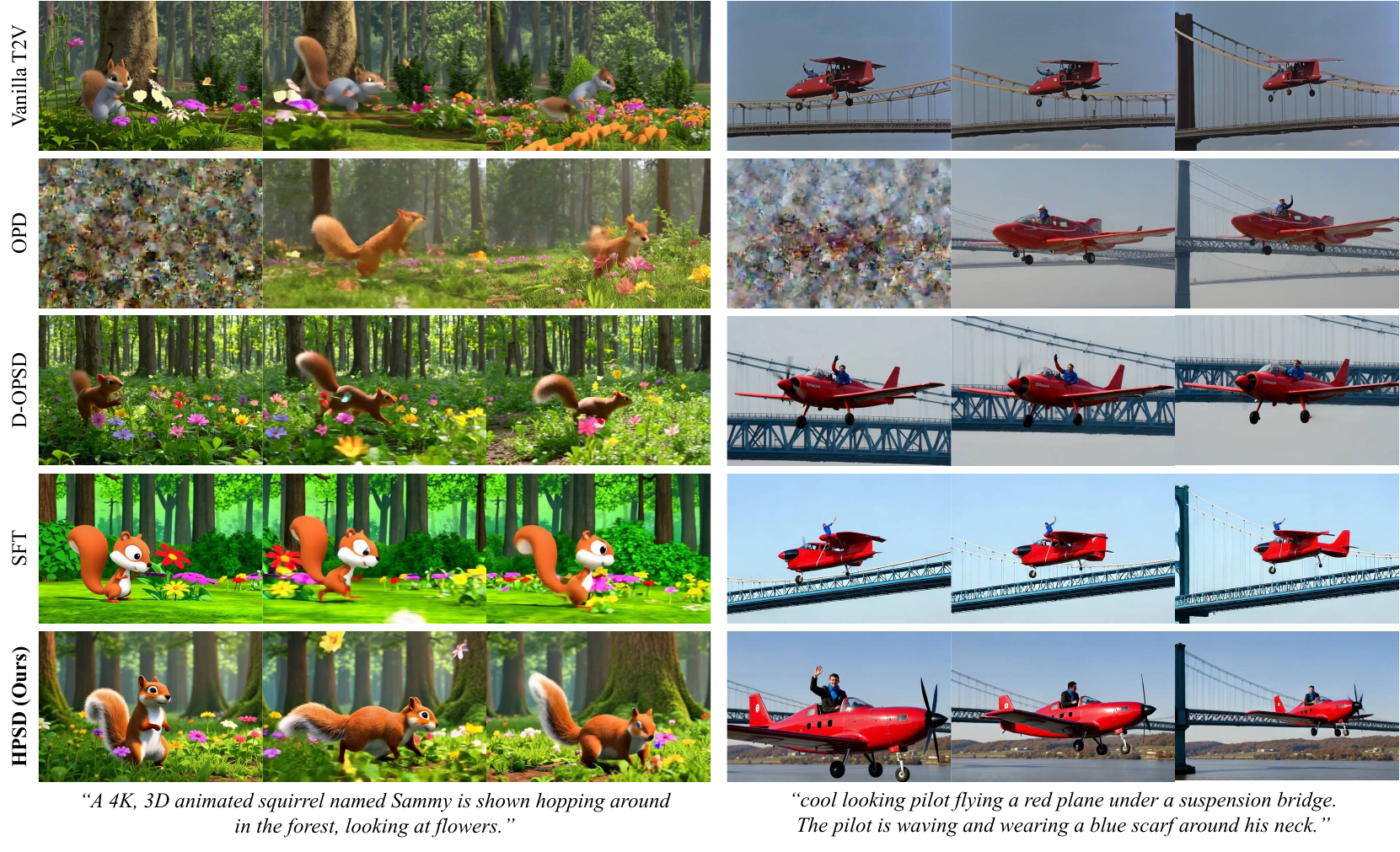}
    \vspace{-1em}
    \caption{
        \textbf{More qualitative comparison results on LTX-2.3 (6/6).} 
        Best viewed zoomed in.
        }
    \label{fig:ltx23-add-6}
\end{figure}

\begin{figure}[t]
    \centering
    \includegraphics[width=1.0\linewidth]{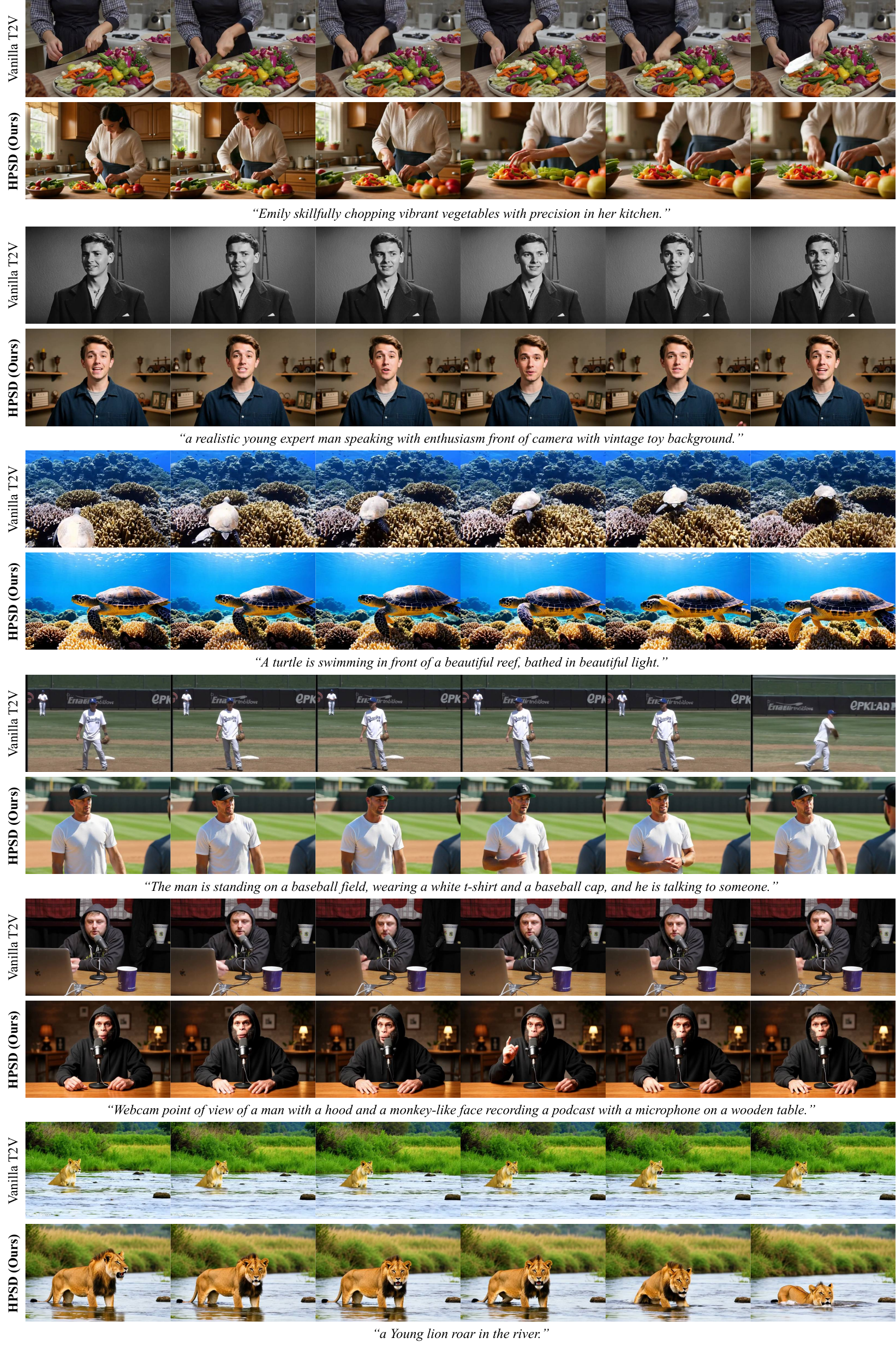}
    \vspace{-1em}
    \caption{
        \textbf{Comparison between HPSD and vanilla LTX-2.3 (1/2).} 
        Best viewed zoomed in.
        }
    \label{fig:ltx23-two-1}
\end{figure}

\begin{figure}[t]
    \centering
    \includegraphics[width=1.0\linewidth]{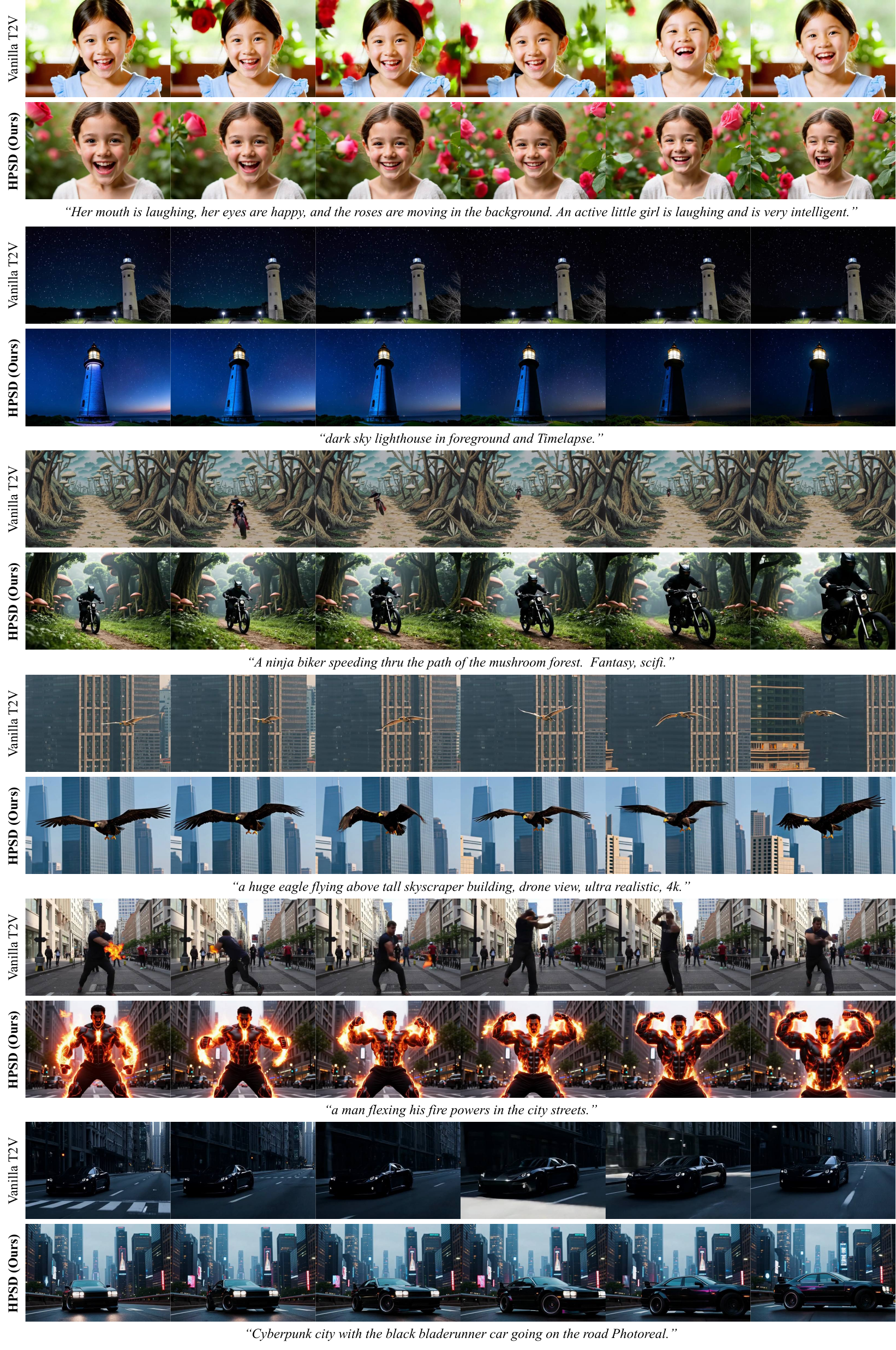}
    \vspace{-1em}
    \caption{
        \textbf{Comparison between HPSD and vanilla LTX-2.3 (2/2).} 
        Best viewed zoomed in.
        }
    \label{fig:ltx23-two-2}
\end{figure}

\begin{figure}[t]
    \centering
    \includegraphics[width=1.0\linewidth]{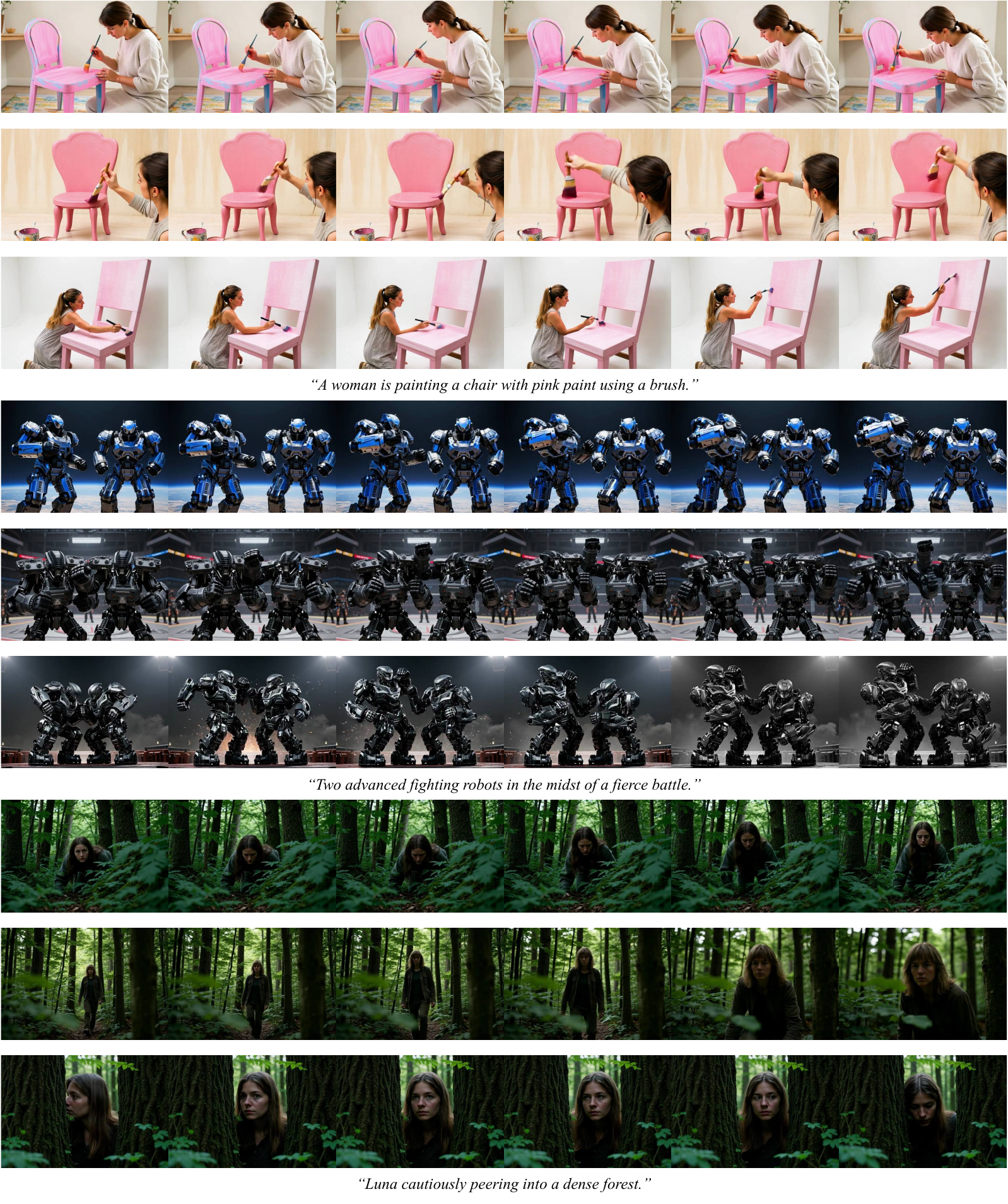}
    \vspace{-1em}
    \caption{
        \textbf{Generated results using same prompts and different seeds on LTX-2.3.} 
        }
    \label{fig:hpsd-seed}
\end{figure}

In this section, we provide further visual results. Specifically, Fig.~\ref{fig:wan22-add-1}-~\ref{fig:wan22-two-3} present additional comparisons between WAN-2.2 and the baselines, while Fig.~\ref{fig:ltx23-add-1}-~\ref{fig:ltx23-two-2} showcase further results for LTX-2.3. 
Videos generated using the same prompts with different random seeds are shown in Fig.~\ref{fig:hpsd-seed}.

\section{Additional Experimental Results}\label{sec:add-quantitative}

In the main experiments, we employed Z-Image-Turbo as the first-frame generator to construct the privileged image conditions. 
To validate the generalizability of HPSD, we conduct an additional study by utilizing a different first-frame generator, specifically Flux.2-Klein-4B~\citep{flux-2-2025}.
As shown in Tab.~\ref{tab:results_flux}, HPSD consistently maintains its performance superiority over the original T2V results, demonstrating that our capability internalization framework is highly robust and generalizes well across various auxiliary text-to-image models.

\begin{figure}[t]
    \centering
    \includegraphics[width=1.0\linewidth]{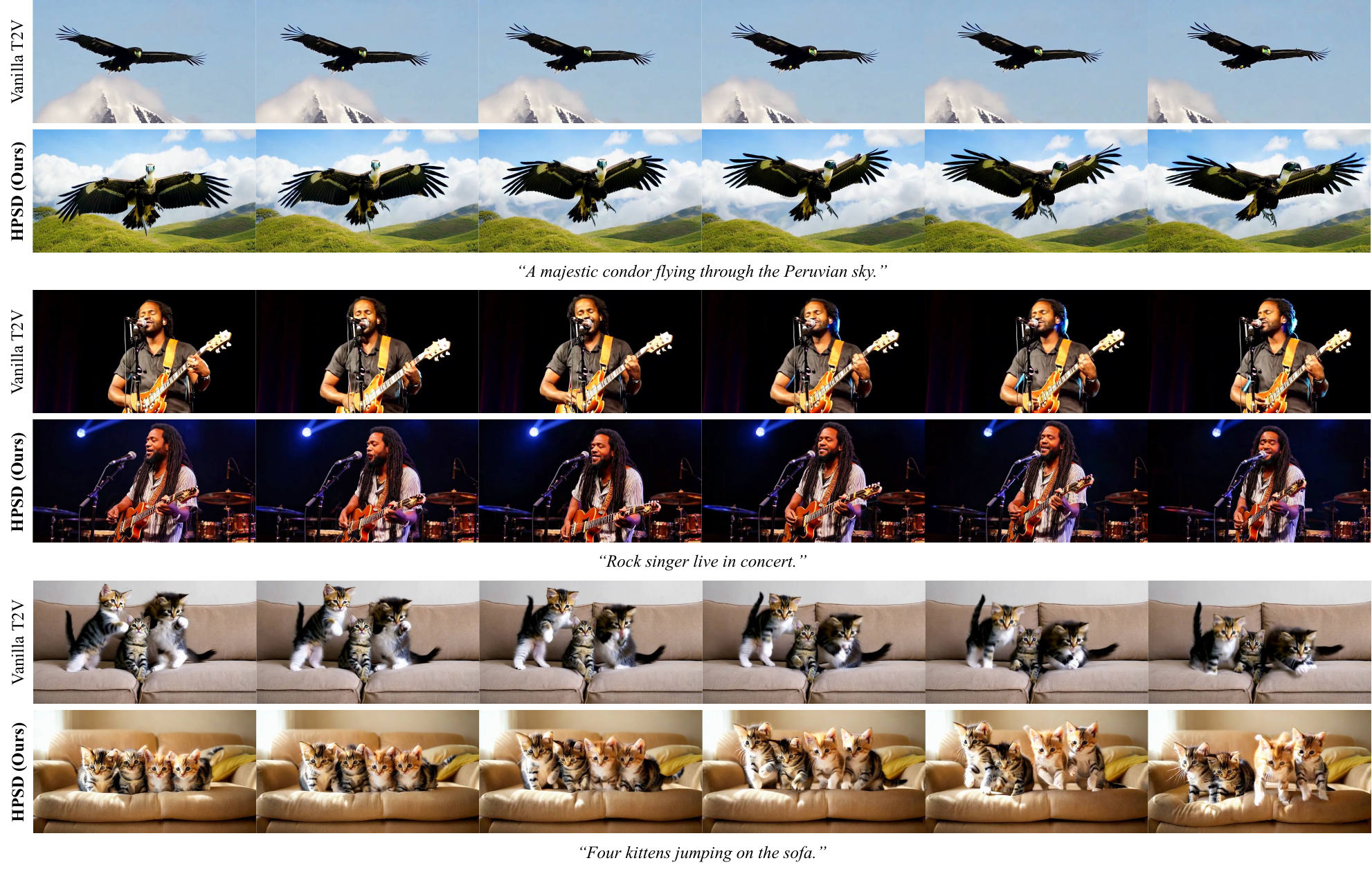}
    \vspace{-1em}
    \caption{
        \textbf{Comparison between HPSD (with Flux.2-Klein-4B) and vanilla WAN-2.2.} 
        }
    \label{fig:hpsd-flux}
\end{figure}

Furthermore, we provide qualitative examples in Fig.~\ref{fig:hpsd-flux}. Visual inspections corroborate the quantitative findings, demonstrating that HPSD consistently delivers high-fidelity, temporally coherent videos across different first-frame generators utilized during the offline construction stage.

\begin{table}[t]
\centering
\caption{
Experimental results using Flux.2-Klein-4B as the first-frame generator.
}
\label{tab:results_flux}
\resizebox{0.9\linewidth}{!}{
\begin{tabular}{lccccccccc}
\toprule
\textbf{Backbone} & \textbf{Method} & \textbf{VideoAlign} & \textbf{VisionReward} & \textbf{UR-v1} & \textbf{UR-v2-A} & \textbf{UR-v2-P} & \textbf{UR-v2-S} & \textbf{HPS} & \textbf{CLIP} \\ \midrule
\multirow{2}{*}{\textbf{WAN-2.2}}
& Vanilla T2V & 0.5335 & 0.0965 & 2.683 & 2.802 & 3.167 & 3.100 & 0.2472 & 0.3684 \\
& \cellcolor{color3}{\textbf{HPSD (Ours)}} & \cellcolor{color3}{\textbf{1.6587}} & \cellcolor{color3}{\textbf{0.1167}} & \cellcolor{color3}{\textbf{2.793}} & \cellcolor{color3}{\textbf{2.865}} & \cellcolor{color3}{\textbf{3.171}} & \cellcolor{color3}{\textbf{3.252}} & \cellcolor{color3}{\textbf{0.2819}} & \cellcolor{color3}{\textbf{0.3785}} \\
\bottomrule
\end{tabular}
}
\end{table}

\section{Text Prompts for Video Generation}\label{sec:prompt}
Text prompts used to generate videos in this paper are provided in Tab.~\ref{tab:prompt1} and Tab.~\ref{tab:prompt2}.

\section{Privileged Condition Examples}\label{sec:condition}
\begin{figure}[t]
    \centering
    \includegraphics[width=0.75\linewidth]{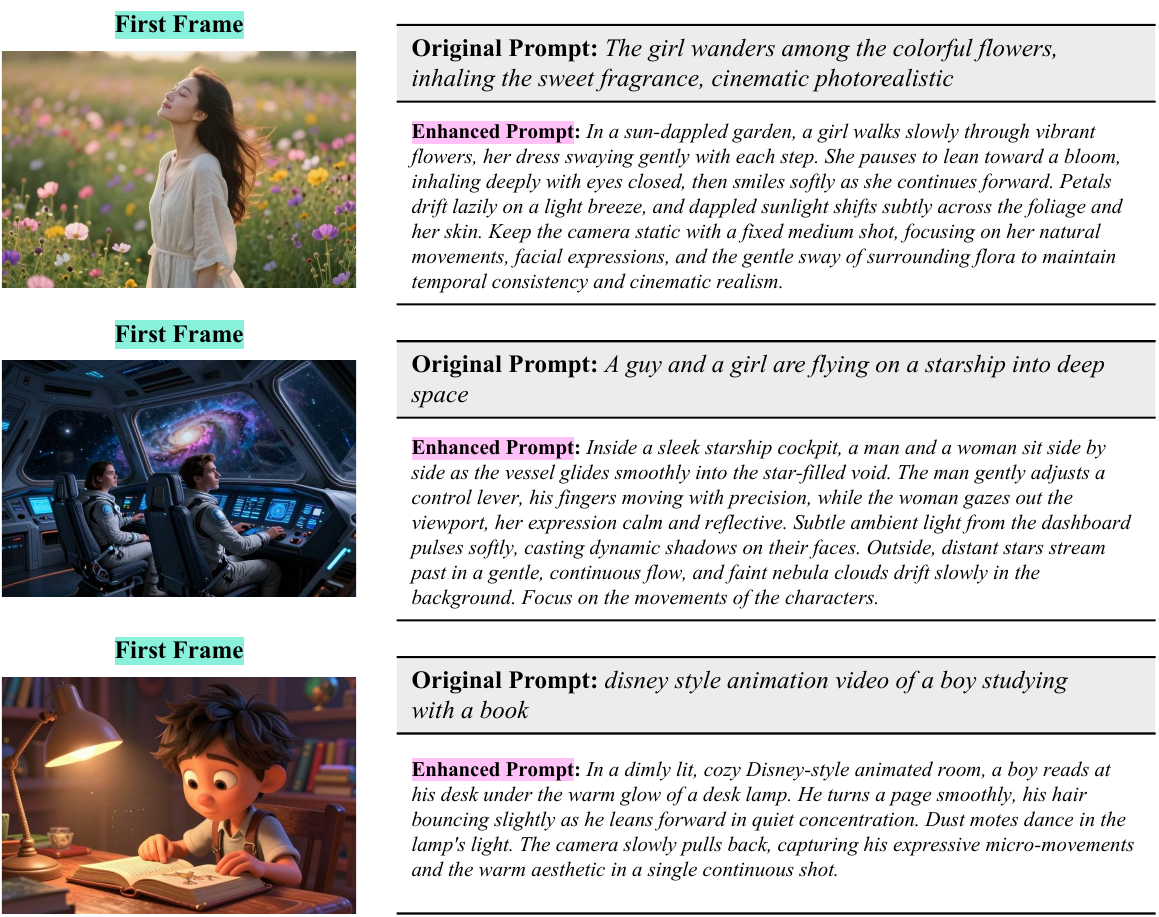}
    \caption{
        \textbf{Privileged Condition Examples.} 
        }
    \label{fig:privileged-condition-example}
\end{figure}

In this section, we present concrete examples of the privileged conditions utilized during the HPSD training process. As depicted in Fig.~\ref{fig:privileged-condition-example}, we showcase the original vanilla text prompts alongside their corresponding enhanced prompts ($c_\text{txt}^{+}$) and synthesized high-quality first frames ($c_\text{img}$). These privileged inputs are leveraged by the teacher to produce the condition-elicited anchor trajectories.

\section{Ethical Statement}\label{sec:ethical}
Throughout the course of this research, we are deeply committed to upholding rigorous ethical standards and fostering the responsible development of generative AI. To the best of our understanding, the proposed distillation framework, along with the utilized datasets and model architectures, does not introduce any novel ethical hazards or societal risks. Furthermore, all experimental procedures and empirical evaluations were conducted in strict compliance with recognized community norms, thereby ensuring the scientific integrity, transparency, and reliability of our findings.

\section{Reproducibility Statement}\label{sec:reproducibility}
In alignment with the principles of transparent research, we strive to make our experimental results seamlessly reproducible for the academic community. To achieve this, the complete source code of HPSD will be fully open-sourced. Furthermore, we have provided comprehensive hyperparameter configurations, prompt designs, and implementation details. We hope this framework will serve as a valuable reference for future studies on diffusion model distillation, inspiring new methodological innovations and driving continuous momentum in the domain.

\section{Declaration on LLM Usage in Writing}\label{sec:declaration}

In this paper, LLMs are utilized only for minor language polishing.

\newpage

\clearpage
\begin{table}[!t]
    \centering
    \caption{The video generation prompts for each figure are listed sequentially, following the order from left to right and top to bottom. (Table 1/2)}
     \label{tab:prompt1}
     \resizebox{\textwidth}{!}{%
        \begin{tabular}{>{\arraybackslash}p{2cm} >{\arraybackslash}p{15cm}}
            \toprule
           \textbf{\centering{Figure}}&\textbf{\centering Text Prompt}\\
    \midrule 
            \multirow{7}{*}{Figure.~\ref{fig:teaser}} &{3D cartoon, a cowboy with an ak47 aiming at you.} \\
             &\cellcolor{color4} cars doing a super race in New York. \\      
              & photo of coastline, rocks, distant lighthouse in the background, storm weather, strong wind, crashing waves, huge water splashes, lightning, 8k uhd, dslr, soft lighting, high quality, film grain, Fujifilm XT3  \\     
               &\cellcolor{color4} The player fights a terrifying skeleton in Minecraft, create this video in a retro video game style. \\ 
               &  A man rides a bike, escaping a lava tsunami, while pigeons fly away in the background and the camera circles him. \\  
   \midrule
            \multirow{1}{*}{Figure.~\ref{fig:observation}} &{purple car in a videogame.} \\

    \midrule
            \multirow{2}{*}{Figure.~\ref{fig:mismatch}} &{A cartoon depicting a penguin with a book balanced on its head, attempting to explain various concepts to a group of rocks.} \\
    \midrule
            \multirow{2}{*}{Figure.~\ref{fig:wan22}} &{cinematic scene of a cute realistic poodle dog in the beach.} \\
            &\cellcolor{color4} A man sitting in cake shop, eating a cake. \\
    \midrule
            \multirow{2}{*}{Figure.~\ref{fig:ltx23}} &{a man playing guitar and sing a song on the street.} \\
            &\cellcolor{color4} a video of a giant blue butterfly of 200 meters in a city, flying ver beutifully. \\
    \midrule
            \multirow{2}{*}{Figure.~\ref{fig:ti2v}} &{A 3D illustration of a man engaging in an MMA fight, rendered in Pixar style.} \\
            &\cellcolor{color4} two dogs wearing goggles and parachute on an airplane. \\
    \midrule
            \multirow{2}{*}{Figure.~\ref{fig:consistency_physics}} &{man in yellow hazmat suit on a paddleboard in the middle of the ocean.} \\
            &\cellcolor{color4} coffee swirling in a rainbow cup. \\
     \midrule
            \multirow{3}{*}{Figure.~\ref{fig:hpsd-flux}} &{A majestic condor flying through the Peruvian sky.} \\
            &\cellcolor{color4} Rock singer live in concert. \\
            & Four kittens jumping on the sofa. \\
    \midrule
            \multirow{2}{*}{Figure.~\ref{fig:wan22-add-1}} &{barbie walking outdoors on a street, perfect proportions.} \\
            &\cellcolor{color4} A bird is walking on a wooden floor and a person is holding a waffle. \\
    \midrule
            \multirow{2}{*}{Figure.~\ref{fig:wan22-add-2}} &{An old wizard with a gray beard, wearing a brown cassock and carrying a staff.} \\
            &\cellcolor{color4} Man in a business suit with headphones walking in the rain at night. \\
    \midrule
            \multirow{2}{*}{Figure.~\ref{fig:wan22-add-3}} &{a ship sailing in the ocean.} \\
            &\cellcolor{color4} A gray Power Ranger, adorned with a purple cape, is running at a high speed against a green background. \\
    \midrule
            \multirow{2}{*}{Figure.~\ref{fig:wan22-add-4}} &{A polar bear operating an iPhone.} \\
            &\cellcolor{color4} Humans walking around a Dragon Zoo 1920s New York, rare film footage. \\
    \midrule
            \multirow{2}{*}{Figure.~\ref{fig:wan22-add-5}} &{steampunk themed chefs cooking beside the sea, high detail rpg.} \\
            &\cellcolor{color4} A boy is picking up a coin from the road in a 3D cartoon. \\
    \midrule
            \multirow{3}{*}{Figure.~\ref{fig:wan22-add-6}} &{A car with wings is flying through a city.} \\
            &\cellcolor{color4} A man rides a bike, escaping a lava tsunami, while pigeons fly away in the background and the camera circles him. \\
    \midrule
            \multirow{6}{*}{Figure.~\ref{fig:wan22-two-1}} &{1950s film, woman doing yoga, standing in warrior 1 pose and move, standing.} \\
            &\cellcolor{color4} boy going school in red color cycle, with a dog. \\
            & Darth Vader drinks whiskey. \\
            &\cellcolor{color4} A robot is walking in the foggy city at sunset. \\
            & A cute puppy running on the grass, sunshine, rainbow, reality. \\
            &\cellcolor{color4} In the 1970s there was an anime about a cute alien girl who shot bubbles from a ray gun at a monster robot. \\
    \bottomrule
        \end{tabular}
    }
\end{table}

\clearpage
\begin{table}[!t]
    \centering
    \caption{The video generation prompts for each figure are listed sequentially, following the order from left to right and top to bottom. (Table 2/2)}
     \label{tab:prompt2}
     \resizebox{\textwidth}{!}{%
        \begin{tabular}{>{\arraybackslash}p{2cm} >{\arraybackslash}p{15cm}}
            \toprule
           \textbf{\centering{Figure}}&\textbf{\centering Text Prompt}\\
    \midrule
            \multirow{6}{*}{Figure.~\ref{fig:wan22-two-2}} &{A red-haired man in a suit and hat is standing next to a parked car, jumping up and down with excitement.} \\
            &\cellcolor{color4} distant view, a line of geese fly across the night moon sky, below the sky is the wild pond. \\
            & Teddy Bear casting as doctor, holding first aid kit bag running, on the sea island. \\
            &\cellcolor{color4} Polaris rzr 1000 flipping down a mountain. \\
            & Children talking in the garden, cartoon drawing. \\
            &\cellcolor{color4} man in yellow hazmat suit on a paddleboard in the middle of the ocean. \\
    \midrule
            \multirow{6}{*}{Figure.~\ref{fig:wan22-two-3}} &{Aphrodite is walking in the forest at sunset.} \\
            &\cellcolor{color4} a beach with blue and green waters, birds flying in the sky and a speedboat. \\
            & people dancing at a sci fi festival with cool robots and a space ship in the sky. \\
            &\cellcolor{color4} a black horse galloping on the sea, full moon, midnight. \\
            & speedly running race cars in a rallye course. \\
            &\cellcolor{color4} faroe islands viking walking on a mauntain looking over the ocean. \\
    \midrule
            \multirow{2}{*}{Figure.~\ref{fig:ltx23-add-1}} &{Madhurima wearing blue leather jacket staring at camera and Jeans, closing eyes slowly.} \\
            &\cellcolor{color4} a green parrot playing a concertina. \\
    \midrule
            \multirow{2}{*}{Figure.~\ref{fig:ltx23-add-2}} &{an alligator swimming in Hagerstown City Park.} \\
            &\cellcolor{color4} a young guy executive on the phone walking across street intersection, lots of people, traffice. \\
    \midrule
            \multirow{2}{*}{Figure.~\ref{fig:ltx23-add-3}} &{People walking on the moon surface, Earth is seen in the sky, Futuristic Homes.} \\
            &\cellcolor{color4} Two teddy bears are talking to each other as they walk into a forest. The scene is depicted in a cartoon style. \\
    \midrule
            \multirow{3}{*}{Figure.~\ref{fig:ltx23-add-4}} &{A red dragon flies in the city.} \\
            &\cellcolor{color4} A deep blue baby turtle, playing the guitar and singing into the microphone, highly detailed, 3 d, dreamy, raytracing. \\
    \midrule
            \multirow{2}{*}{Figure.~\ref{fig:ltx23-add-5}} &{Animation of drifting waves, with moving pink clouds at sunset.} \\
            &\cellcolor{color4} A black Labrador and a gray Norwegian Forest cat play on the lawn in the backyard. \\
    \midrule
            \multirow{3}{*}{Figure.~\ref{fig:ltx23-add-6}} &{A 4K, 3D animated squirrel named Sammy is shown hopping around in the forest, looking at flowers.} \\
            &\cellcolor{color4} cool looking pilot flying a red plane under a suspension bridge. The pilot is waving and wearing a blue scarf around his neck. \\
    \midrule
            \multirow{8}{*}{Figure.~\ref{fig:ltx23-two-1}} &{Emily skillfully chopping vibrant vegetables with precision in her kitchen.} \\
            &\cellcolor{color4} a realistic young expert man speaking with enthusiasm front of camera with vintage toy background. \\
            & A turtle is swimming in front of a beautiful reef, bathed in beautiful light. \\
            &\cellcolor{color4} The man is standing on a baseball field, wearing a white t-shirt and a baseball cap, and he is talking to someone. \\
            & Webcam point of view of a man with a hood and a monkey-like face recording a podcast with a microphone on a wooden table. \\
            &\cellcolor{color4} a Young lion roar in the river. \\
    \midrule
            \multirow{7}{*}{Figure.~\ref{fig:ltx23-two-2}} &{Her mouth is laughing, her eyes are happy, and the roses are moving in the background. An active little girl is laughing and is very intelligent.} \\
            &\cellcolor{color4} dark sky lighthouse in foreground and Timelapse. \\
            & A ninja biker speeding thru the path of the mushroom forest. Fantasy, scifi. \\
            &\cellcolor{color4} a huge eagle flying above tall skyscraper building, drone view, ultra realistic, 4k. \\
            & a man flexing his fire powers in the city streets. \\
            &\cellcolor{color4} Cyberpunk city with the black bladerunner car going on the road Photoreal. \\
    \midrule
            \multirow{3}{*}{Figure.~\ref{fig:hpsd-seed}} &{A woman is painting a chair with pink paint using a brush.} \\
            &\cellcolor{color4} Two advanced fighting robots in the midst of a fierce battle. \\
            & Luna cautiously peering into a dense forest. \\
    \bottomrule
        \end{tabular}
    }
\end{table}

\end{document}